\PassOptionsToPackage{type1,nofontspec}{XCharter}
\documentclass[11pt,a4paper,copyright]{googlecloud}
\microtypesetup{tracking=false}

\usepackage[authoryear,sort&compress,round]{natbib}
\hypersetup{colorlinks=true,citecolor=blue,linkcolor=red,urlcolor=blue,
  pdftitle={GPRM: Global Perspective Process Reward Model},
  pdfauthor={Zifan Zhang, Zhengbo Jiao, Shaobo Wang, Jing Zhou, Guoyi Li, Jiayi Chen, Wei Wang, Bing Zhao, Cheng Fang, Xiaoxiao Xu, Hu Wei, Linfeng Zhang, Lin Qu}}
\usepackage{subcaption}
\usepackage{multirow}
\usepackage{makecell}
\usepackage{float}
\usepackage{placeins}
\newcommand{\down}[1]{\textcolor{red!70!black}{\scriptsize $\downarrow$#1}}
\renewcommand{\today}{November 2026}
\renewcommand{\copyrightext}{\footerfont * Equal contribution.\quad $\dagger$ Corresponding author.}
\title{GPRM: Global Perspective Process Reward Model}
\author[1,7*]{Zifan Zhang}
\author[2,7*]{Zhengbo Jiao}
\author[3]{Shaobo Wang}
\author[4]{Jing Zhou}
\author[5]{Guoyi Li}
\author[6]{Jiayi Chen}
\author[7]{Wei Wang}
\author[7]{Bing~Zhao}
\author[7]{Cheng Fang}
\author[7]{Xiaoxiao Xu}
\author[7]{Hu Wei}
\author[3$\dagger$]{Linfeng Zhang}
\author[7]{Lin Qu}
\affil[1]{Wuhan University}
\affil[2]{Shanghai University of Finance and Economics}
\affil[3]{Shanghai Jiao Tong University}
\affil[4]{Fudan University}
\affil[5]{University of Chinese Academy of Sciences}
\affil[6]{Beijing University of Posts and Telecommunications}
\affil[7]{Alibaba Group}
\correspondingauthor{linfeng.zhang@sjtu.edu.cn}

\begin{abstract}
Process Reward Models (PRMs) are widely used for step-level evaluation of multi-step reasoning in large language models, but existing methods are often limited by weak historical dependency modeling and insufficient global awareness. They typically fail to leverage prior step evaluations for coherent judgment and lack visibility into subsequent reasoning, leading to suboptimal assessment of intermediate steps. To address this issue, we propose the Global Perspective Process Reward Model (GPRM), which jointly models the problem, previous reasoning steps and their evaluations, the current step, and future steps to enable history-aware and globally informed process evaluation. We further introduce a structured evaluation framework along four dimensions---look back, look ahead, self-check, and goal alignment---to assess prior consistency, future plausibility, step validity, and task alignment. Based on this framework, we construct datasets for supervised fine-tuning (SFT) and reinforcement learning (GRPO). Experiments show that, despite using less training data and smaller model sizes, GPRM outperforms strong baseline models by 7.8 and 5.4 points on PRMBench and ProcessBench, respectively. GPRM also generalizes well across tasks, surpassing the GPT-4.1 baseline on Agent Error Bench, improving strong baselines by around 4 points under test-time scaling (TTS), and achieving competitive performance on PPE.
\end{abstract}

\begin{document}
\maketitle
\begin{figure}[H]
    \centering
    \captionsetup{font=small,skip=6pt}
    \includegraphics[width=\columnwidth]{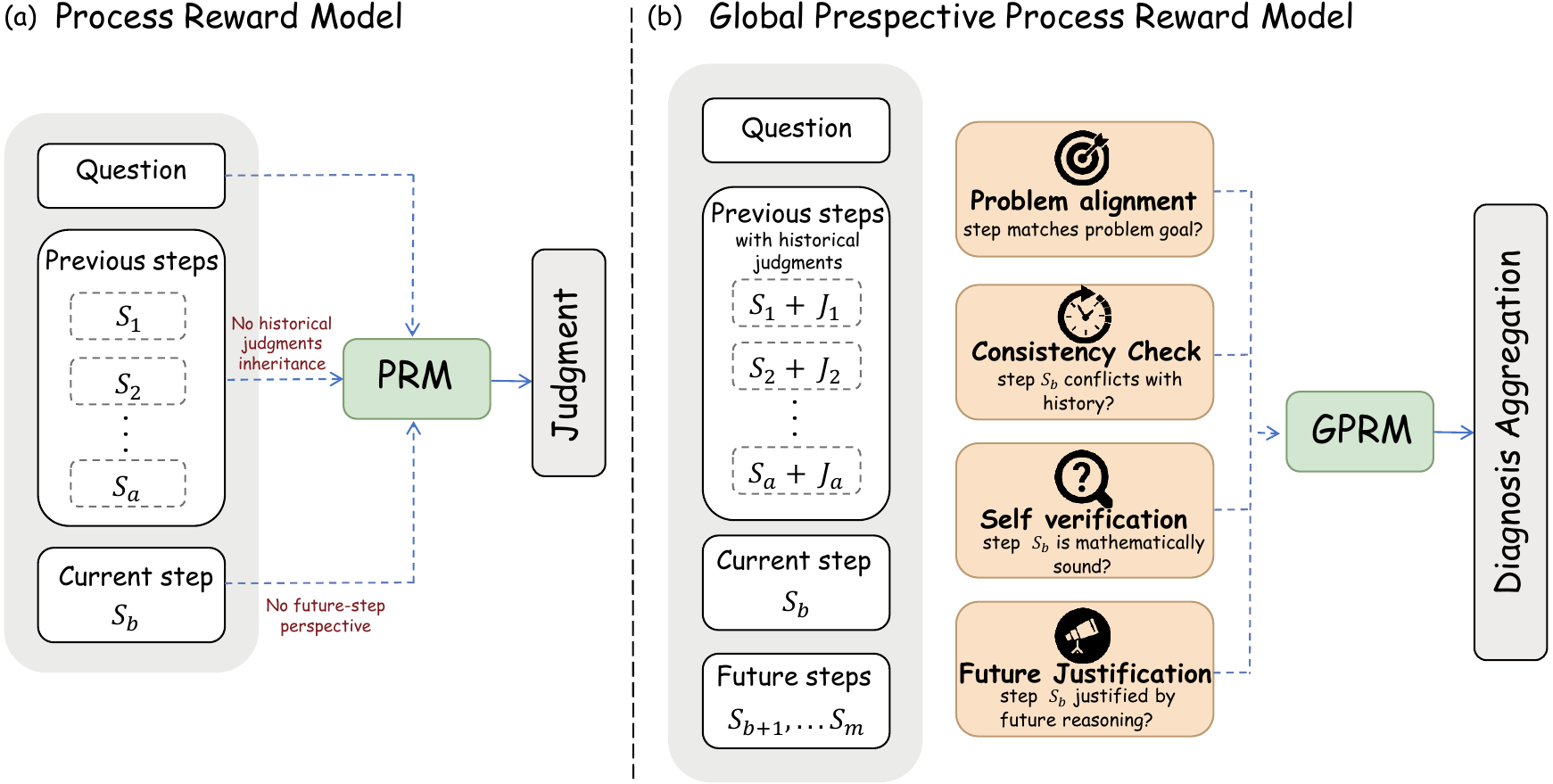}
    \caption{(a) Conventional PRMs judge the current step from local context only, without explicit inheritance of historical judgments or future-step perspective. (b) GPRM evaluates each step under a history--current--future view, combining historical judgments and future reasoning, and aggregates four diagnostic signals---Look-back, Look-ahead, Self-check, and Problem-alignment---into the final judgment.}

    \label{fig:gprm_framework}
\end{figure}

\section{Introduction}

Large language models (LLMs) have achieved remarkable progress on multi-step tasks such as mathematical reasoning, code generation, and complex decision making, with further gains from chain-of-thought (CoT) prompting and test-time search \citep{shao2024deepseekmathpushinglimitsmathematical,yang2024qwen25mathtechnicalreportmathematical}. However, LLMs still frequently make mistakes in intermediate reasoning steps, including logical errors, calculation mistakes, missing conditions, and deviations from the original objective. Compared with errors in the final answer, these process-level errors are often more subtle and more likely to be propagated and amplified by later steps, ultimately causing the entire reasoning trajectory to drift away from the correct solution. To address this issue, Process Reward Models (PRMs) have been introduced to provide fine-grained evaluation of intermediate steps. Unlike Outcome Reward Models (ORMs), which focus only on the final answer, PRMs can localize where errors occur and provide more precise supervision signals for model training \citep{xiong2025stepwiserstepwisegenerativejudges,ye2025correctnessharmonizingprocessoutcome}, inference-time search, and error correction. As a result, PRMs have become a key component for improving the robustness of multi-step reasoning and have been widely used in mathematical verification, test-time search \citep{liu20251bllmsurpass405b}, and process supervision \citep{she2025rprmreasoningdrivenprocessreward,wang2024mathshepherdverifyreinforcellms}.

\begin{figure}[t]
    \centering
    \includegraphics[width=\columnwidth]{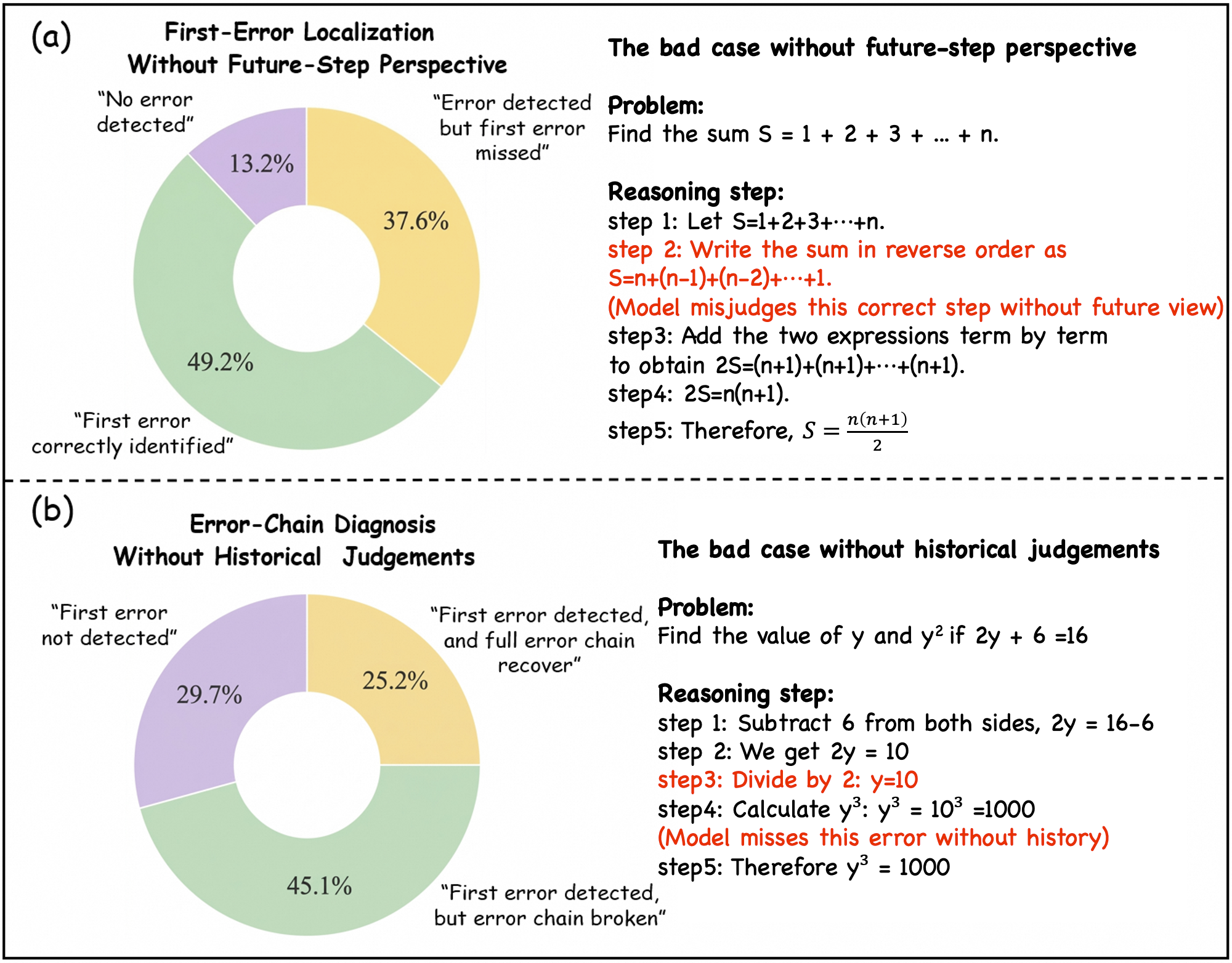}
    \caption{Two limitations of existing PRMs and the effect of adding the future-step perspective and historical judgment inheritance. (a)~Without the future-step perspective, only 49.2\% of cases locate the first error correctly (37.6\% mis-localize, 13.2\% miss); adding future steps cuts mis-localization from 37.6\% to 15.7\%, largely because ``aha-moment'' detours are no longer mispenalized. (b)~Without historical judgment inheritance, 45.1\% find the first error but break the chain and 29.7\% miss it entirely; inheriting prior judgments cuts the broken-chain rate from 45.1\% to 35.5\% by keeping a coherent view of all preceding steps.}
    \label{fig:case}
\end{figure}

Despite recent progress, existing PRMs still suffer from two major limitations. First, they lack historical judgment inheritance: most methods evaluate the current step based only on the question, preceding step texts, and the current step itself, without explicitly incorporating judgments made on earlier steps. As a result, although the model can observe the history \citep{zhao2025genprmscalingtesttimecompute}, it often fails to consistently leverage previously established diagnoses across long reasoning chains. Second, they lack a global perspective~\citep{zou2025reasonfluxprmtrajectoryawareprmslong}: existing PRMs typically assess each step from local context alone, without considering how subsequent reasoning unfolds. However, when performing complex reasoning (e.g., test-time compute scaling), large language models (LLMs) often utilize reflection, backtracking, or experience ''aha-moments''. During these ''aha-moments'', seemingly flawed or highly exploratory steps are actually necessary turning points that eventually lead to correct reasoning. A local, linear PRM mistakes such a detour for the first error, whereas a future-step perspective can see that the step is taken up and validated by later reasoning. Figure~\ref{fig:case} quantifies both limitations.

To address these challenges, we propose the Global Perspective Process Reward Model (GPRM). As shown in Figure~\ref{fig:gprm_framework}, GPRM evaluates a target step by jointly conditioning on the question, the preceding steps together with their historical judgments, the current step, and the future steps. Unlike conventional PRMs that score each step only from local context, this design explicitly models dependencies across the \textit{history--current--future} structure, enabling both historical inheritance and global diagnosis. In addition, we introduce a unified generative four-dimensional diagnostic framework for process-level errors: Look-back, Look-ahead, Self-check, and Problem-alignment. These dimensions respectively examine consistency, justification from subsequent reasoning, intrinsic logical and computational validity, and alignment with the original task objective. Under this framework, step evaluation is formulated not as an isolated local classification problem, but as a structured diagnostic task over the full trajectory.

To equip the model with these capabilities, we further design a training pipeline tailored to GPRM. In the supervised fine-tuning stage, we construct high-quality mathematical reasoning data under the four diagnostic dimensions and use a strong teacher model to generate step-level analyses, enabling the model to learn structured diagnosis and explanation. On top of this, we mine hard examples from PRM800K, focusing on low-confidence, long-horizon, and late-error cases, and build GRPO training instances to further improve robustness in challenging reasoning scenarios.

Extensive experiments demonstrate the effectiveness of GPRM in process supervision, cross-task generalization, and test-time search. On PRMBench and ProcessBench, GPRM surpasses mainstream GPT-4o and Gemini-2.0-thinking with a smaller model scale. On Agent Error Bench, it generalizes from mathematical reasoning to agent trajectory diagnosis without task-specific adaptation, and surpasses GPT-4.1. Under test-time search settings, GPRM further improves strong baselines. These results highlight the importance of global perspective and historical judgment inheritance for process-level reasoning evaluation. Our contributions can be summarized as follows:
\begin{enumerate}[leftmargin=*,topsep=0.5pt,itemsep=0pt,parsep=0pt]
    \item \textbf{Identifying Key Limitations.} We identify two critical weaknesses in existing PRMs---lack of historical judgment inheritance and absence of global perspective---and demonstrate through quantitative analysis (Figure~\ref{fig:case}) that these lead to locally plausible yet globally incorrect judgments in long-chain reasoning.
    
    \item \textbf{GPRM Framework.} We propose GPRM, which evaluates each step by jointly conditioning on the question, historical judgments $\mathcal{H}_i$, current step, and future trajectory $\mathcal{F}_i$ (Figure~\ref{fig:gprm_framework}). We introduce a four-dimensional diagnostic framework (Look-back, Look-ahead, Self-check, Problem-alignment) and a progressive training pipeline combining SFT with hard-example-based GRPO.
    
    \item \textbf{Superior Performance.} GPRM-4B achieves 73.9 on PRMBench and 74.3 on ProcessBench, outperforming 7B PRMs and proprietary models including GPT-4o (66.8, 61.9) and o1-mini (68.8). GPRM generalizes to agent error localization (54.0\%) and improves test-time search by +5.6 points under Greedy Guided Search@8.
\end{enumerate}

\begin{figure}[t]
    \centering
    \includegraphics[width=\textwidth]{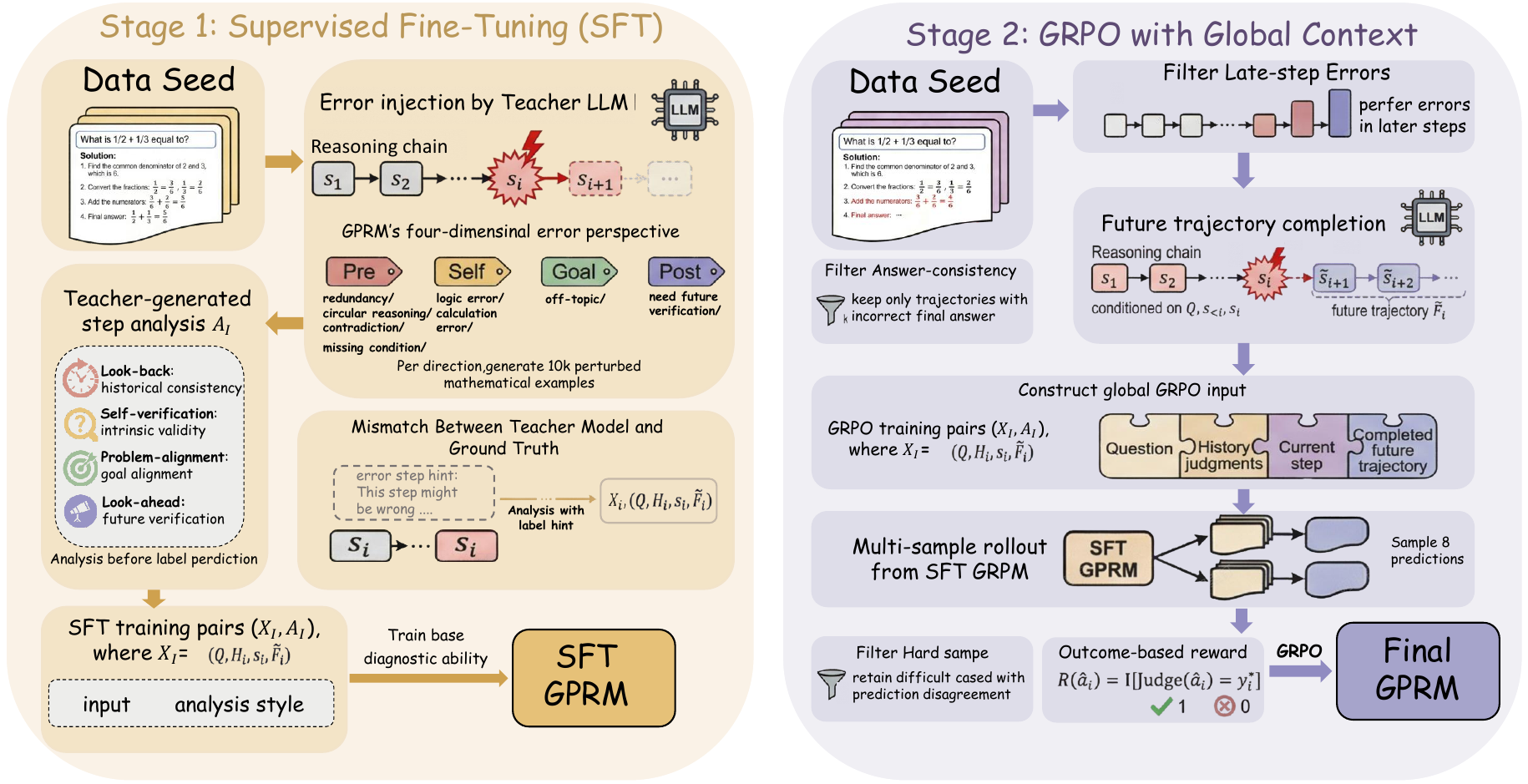}

    \caption{Overall training framework of GPRM. Stage I performs supervised fine-tuning on teacher-constructed step-analysis data with injected errors to learn four-dimensional diagnosis. Stage II introduces global context by completing future trajectories, constructing history-aware inputs, mining hard samples, and optimizing the model with GRPO.}

    \label{fig:gprmf}
\end{figure}

\section{Related Work}

\paragraph{Process Reward Models.}
Process reward models (PRMs) provide step-level supervision for multi-step reasoning, complementing outcome reward models~\citep{liu20251bllmsurpass405b,park2024ensemblinglargelanguagemodels,chen2026rmr1rewardmodelingreasoning,guo2025rewardreasoningmodel}. High-quality supervision is costly, so methods such as Monte-Carlo estimation, rule-based construction, and automatic annotation have been developed~\citep{wang2024mathshepherdverifyreinforcellms,luo2024improvemathematicalreasoninglanguage,lightman2023letsverifystepstep}, with later work improving PRMs via LLM-as-a-judge, preference optimization, retrieval, and dynamic verification~\citep{zhang2025lessonsdevelopingprocessreward}.

\paragraph{Discriminative vs.\ Generative PRMs.}
Early PRMs predict a correctness label or scalar reward per step~\citep{zou2025reasonfluxprmtrajectoryawareprmslong,yang2024qwen25mathtechnicalreportmathematical,wang2024mathshepherdverifyreinforcellms,luo2024improvemathematicalreasoninglanguage,setlur2024rewardingprogressscalingautomated,li2025processrewardmodelqvalue}, underutilizing LLM reasoning. Generative PRMs instead analyze the step before judging: R-PRM~\citep{she2025rprmreasoningdrivenprocessreward}, GenPRM~\citep{zhao2025genprmscalingtesttimecompute}, and ThinkPRM~\citep{khalifa2025processrewardmodelsthink} each reason over the current step, but from scratch and without inheriting prior judgments.

\paragraph{Global and Future-Step Perspective.}
Several works use future or global information, but collapse it into a scalar rather than reasoning over the future text: BiPRM~\citep{zhang2026bidirectionalprocessrewardmodel} re-scores each step against following steps in a right-to-left pass, value/advantage-style PRMs let a step's score depend on future success probability from rollout~\citep{wang2024mathshepherdverifyreinforcellms,luo2024improvemathematicalreasoninglanguage}, and CRM~\citep{zhang2026linkingprocessoutcomeconditional} links step rewards to the final outcome via the chain rule of conditional probabilities. R-PRM analyzes consistency with preceding steps, yet inspects their \emph{content} rather than prior \emph{judgments}, and no existing generative PRM conditions on the future trajectory text.  PRMs also serve as online verifiers in test-time search~\citep{liu20251bllmsurpass405b,she2025rprmreasoningdrivenprocessreward,zhao2025genprmscalingtesttimecompute}, where this richer view is especially valuable.

\section{Method}\label{sec:method}

\subsection{Problem Formulation}\label{subsec:problem_formulation}

We study the \emph{step-level evaluation} problem in multi-step reasoning 
with large language models. Given a question $Q$ and its corresponding 
reasoning trajectory $\mathbf{s}=(s_1,s_2,\ldots,s_n)$, where $s_i$ denotes 
the $i$-th reasoning step, our goal is to predict a correctness label for 
each step: $y_i \in \{0,1\}$, where $y_i=1$ indicates correctness and 
$y_i=0$ indicates an error.

\begin{table}[tbp]
\centering
\caption{Overall performance of GPRM on process-supervision benchmarks. Subtable (a) reports results on PRMBench, and subtable (b) reports results on ProcessBench. Green rows indicate model categories, pink rows highlight our models, and $\star$ marks the primary model. The last two rows of each subtable compare the SFT and GRPO stages of GPRM-4B: GRPO consistently improves over SFT on both benchmarks (+1.0 on PRMBench, +2.0 on ProcessBench), showing that the gains from the four-dimensional global framework are preserved---and amplified---after reinforcement learning.}

\label{tab:process_supervision}

\definecolor{headergray}{RGB}{242,242,242}
\definecolor{groupgreen}{RGB}{235,243,234}
\definecolor{ourspink}{RGB}{248,228,228}

\fontsize{8}{9}\selectfont
\setlength{\tabcolsep}{2pt}
\renewcommand{\arraystretch}{1.00}

\begin{subtable}{\textwidth}
\centering
\caption{Model performance comparison on PRMBench.}

\resizebox{\linewidth}{!}{%
\begin{tabular}{l ccc ccccc cccc c}
\toprule
\rowcolor{headergray}
& \multicolumn{3}{c}{\textbf{Simplicity}} 
& \multicolumn{5}{c}{\textbf{Soundness}} 
& \multicolumn{4}{c}{\textbf{Sensitivity}} 
& \\
\cmidrule(lr){2-4}\cmidrule(lr){5-9}\cmidrule(lr){10-13}
\rowcolor{headergray}
\textbf{Model Name} & \textbf{NR.} & \textbf{NCL.} & \textbf{Avg.} 
& \textbf{ES} & \textbf{SC.} & \textbf{DC.} & \textbf{CI} & \textbf{Avg.} 
& \textbf{PS} & \textbf{DR.} & \textbf{MS.} & \textbf{Avg.} 
& \textbf{Overall} \\
\midrule

\rowcolor{groupgreen}
\multicolumn{14}{l}{\textit{LLM-as-judge, Proprietary language models}} \\
GPT-4o     & 57.0 & 62.4 & 59.7 & 72.0 & 69.7 & 70.7 & 71.1 & 70.9 & 62.5 & 65.7 & 99.2 & 75.8 & 66.8 \\
o1-mini    & 65.6 & 63.7 & 64.6 & 74.5 & 67.7 & 73.8 & 72.3 & 72.1 & 61.8 & 64.8 & 100.0 & 75.5 & 68.8 \\
Gemini-2.0-flash-exp & 66.0 & 67.2 & 58.1 & 62.7 & 70.4 & 65.7 & 66.0 & 67.3 & 61.8 & 66.2 & 98.2 & 75.4 & 66.9 \\
Gemini-2.0-thinking-exp-1219 & 68.8 & 68.5 & 63.8 & 66.2 & 72.9 & 71.3 & 71.0 & 71.8 & 60.3 & 65.7 & 99.8 & 75.3 & 70.3 \\

\midrule
\rowcolor{groupgreen}
\multicolumn{14}{l}{\textit{LLM-as-judge, Open-source language models}} \\
Qwen2.5-Math-72B-Instruct\citep{yang2024qwen25mathtechnicalreportmathematical}  & 55.3 & 54.9 & 55.1 & 55.5 & 71.6 & 58.1 & 59.1 & 61.1 & 47.4 & 53.8 & 100.0 & 67.1 & 57.4 \\
Qwen2.5-72B-Instruct\citep{qwen2025qwen25technicalreport}       & 57.6 & 54.9 & 56.3 & 61.4 & 65.5 & 65.8 & 61.1 & 63.4 & 48.8 & 54.1 & 100.0 & 67.6 & 57.5 \\
QwQ-32B-Preview\citep{qwq-32b-preview}            & 57.2 & 55.6 & 56.4 & 67.4 & 72.3 & 66.2 & 66.9 & 68.2 & 57.8 & 62.7 & 100.0 & 73.5 & 63.6 \\

\midrule
\rowcolor{groupgreen}
\multicolumn{14}{l}{\textit{PRMs}} \\
Math-Shepherd-7B\citep{wang2024mathshepherdverifyreinforcellms}        & 44.0 & 50.3 & 47.1 & 49.4 & 44.5 & 41.3 & 47.7 & 45.7 & 47.2 & 48.6 & 86.1 & 60.7 & 47.0 \\
RLHFlow-Mistral-8B\citep{xiong2024rlhflowmath}      & 46.1 & 47.3 & 46.7 & 56.6 & 55.1 & 54.4 & 63.8 & 57.5 & 51.5 & 56.2 & 97.9 & 68.5 & 54.4 \\
RLHFlow-DeepSeek-8B\citep{xiong2024rlhflowmath}     & 46.4 & 48.9 & 47.6 & 55.7 & 55.0 & 53.2 & 66.2 & 57.5 & 49.0 & 55.4 & 99.8 & 68.1 & 54.2 \\
Llemma-PRM800K-7B\citep{gpt_accelera}       & 49.3 & 53.4 & 51.4 & 56.4 & 47.1 & 46.7 & 53.3 & 50.9 & 51.0 & 53.5 & 93.6 & 66.0 & 52.0 \\
Skywork-PRM-1.5B \citep{he_2024_16998085}       & 52.0 & 56.4 & 54.2 & 64.8 & 64.9 & 63.3 & 65.5 & 64.9 & 57.5 & 63.3 & 91.1 & 70.7 & 61.1 \\
Skywork-PRM-7B\citep{he_2024_16998085}      & 56.4 & 62.8 & 59.6 & 69.4 & 67.1 & 67.7 & 69.9 & 68.5 & 60.9 & 65.8 & 93.2 & 73.3 & 65.1 \\
Qwen2.5-Math-7B-PRM800K\citep{prmlessons} & 48.6 & 47.8 & 48.2 & 62.1 & 59.4 & 58.7 & 68.5 & 62.2 & 52.9 & 64.0 & 99.8 & 72.2 & 58.3 \\
Qwen2.5-Math-PRM-7B\citep{qwen2025qwen25technicalreport}     & 49.0 & 55.1 & 52.1 & 71.8 & 67.3 & 66.3 & 78.5 & 71.0 & 57.6 & 69.1 & 99.7 & 75.5 & 65.5 \\
R-PRM-7B-SFT\citep{she2025rprmreasoningdrivenprocessreward}            & 52.7 & 64.7 & 58.7 & 70.1 & 62.7 & 63.4 & 69.5 & 66.4 & 61.4 & 67.4 & 98.3 & 75.7 & 64.9 \\
R-PRM-7B-DPO \citep{she2025rprmreasoningdrivenprocessreward}       & 52.2 & 58.2 & 55.2 & 72.1 & 69.1 & 68.9 & 75.0 & 71.2 & 61.2 & 69.5 & 99.1 & 76.6 & 66.8 \\
GenPRM-7B \citep{zhao2025genprmscalingtesttimecompute}              & 53.0 & 59.1 & 56.1 & 72.6 & 69.8 & 69.3 & 75.6 & 71.8 & 61.7 & 70.0 & 99.2 & 77.0 & 67.4 \\

\rowcolor{ourspink}
GPRM-4B-SFT   & 67.0 & 62.9 & 65.0 & 78.2 & 74.3 & 73.5 & 80.8 & 75.2 & 63.8 & 73.5 & 99.2 & 78.8 & 72.9 \\
\rowcolor{ourspink}
GPRM-4B-GRPO  & 68.2 & 63.4 & 65.8 & 79.1 & 75.0 & 74.2 & 81.5 & 76.2 & 64.5 & 74.1 & 99.3 & 79.3 & 73.9 \\
\rowcolor{ourspink}
$\star$ \textbf{GPRM-14B-GRPO} & \textbf{69.5} & \textbf{64.8} & \textbf{67.2} & \textbf{80.3} & \textbf{76.1} & \textbf{75.4} & \textbf{82.6} & \textbf{77.6} & \textbf{65.9} & \textbf{75.2} & 99.5 & \textbf{80.2} & \textbf{74.6} \\
\bottomrule
\end{tabular}%
}
\end{subtable}

\vspace{0.2cm}

\begin{subtable}{\textwidth}
\centering
\caption{Model performance comparison on ProcessBench.}

\resizebox{\linewidth}{!}{%
\begin{tabular}{l ccc ccc ccc ccc c}
\toprule
\rowcolor{headergray}
\multirow{2}{*}{\textbf{Model}} 
& \multicolumn{3}{c}{\textbf{GSM8K}} 
& \multicolumn{3}{c}{\textbf{MATH}} 
& \multicolumn{3}{c}{\textbf{OLYMPIADBENCH}} 
& \multicolumn{3}{c}{\textbf{OMNIMATH}} 
& \multirow{2}{*}{\textbf{Avg.\ F1}} \\
\cmidrule(lr){2-4}\cmidrule(lr){5-7}\cmidrule(lr){8-10}\cmidrule(lr){11-13}
\rowcolor{headergray}
& \textbf{Err.} & \textbf{Cor.} & \textbf{F1}
& \textbf{Err.} & \textbf{Cor.} & \textbf{F1}
& \textbf{Err.} & \textbf{Cor.} & \textbf{F1}
& \textbf{Err.} & \textbf{Cor.} & \textbf{F1}
& \\
\midrule

\rowcolor{groupgreen}
\multicolumn{14}{l}{\textit{LLM-as-judge, Proprietary language models}} \\
GPT-4o*  & 70.0 & 91.2 & 79.2 & 54.4 & 76.6 & 63.6 & 45.8 & 58.4 & 51.4 & 45.2 & 65.6 & 53.5 & 61.9 \\
o1-mini* & 88.9 & 97.9 & 93.2 & 83.5 & 95.1 & 88.9 & 80.2 & 95.6 & 87.2 & 74.8 & 91.7 & 82.4 & \textbf{87.9} \\

\midrule
\rowcolor{groupgreen}
\multicolumn{14}{l}{\textit{LLM-as-judge, Open-source language models}} \\
Llama-3.3-70B-Instruct*\citep{grattafiori2024llama3herdmodels}     & 72.5 & 96.9 & 82.9 & 43.3 & 83.2 & 59.4 & 31.0 & 94.1 & 46.7 & 28.2 & 90.5 & 43.0 & 58.0 \\
Qwen2.5-Math-72B-Instruct*\citep{yang2024qwen25mathtechnicalreportmathematical}  & 49.8 & 96.9 & 65.8 & 36.0 & 94.3 & 52.1 & 19.5 & 97.3 & 32.5 & 19.0 & 96.3 & 31.7 & 45.5 \\
Qwen2.5-72B-Instruct*\citep{qwen2025qwen25technicalreport}       & 62.8 & 96.9 & 76.2 & 46.3 & 93.1 & 61.8 & 38.7 & 92.6 & 54.6 & 36.6 & 90.9 & 52.2 & 61.2 \\
QwQ-32B-Preview\citep{qwq-32b-preview}             & 81.6 & 95.3 & 88.0 & 78.1 & 79.3 & 78.7 & 61.4 & 54.6 & 57.8 & 55.7 & 68.0 & 61.3 & 71.5 \\

\midrule
\rowcolor{groupgreen}
\multicolumn{14}{l}{\textit{PRMs}} \\

Math-Shepherd-7B\citep{wang2024mathshepherdverifyreinforcellms}            & 32.4 & 91.7 & 47.9 & 18.0 & 82.0 & 29.5 & 15.0 & 71.1 & 24.8 & 14.2 & 73.0 & 23.8 & 31.5 \\
Math-PSA-7B\citep{he_2024_16998085}                 & 48.3 & 88.1 & 62.4 & 29.5 & 72.7 & 41.9 & 20.7 & 65.8 & 31.5 & 15.4 & 68.9 & 25.2 & 40.3 \\
RLHFlow-DeepSeek-8B\citep{xiong2024rlhflowmath}         & 24.2 & 98.4 & 38.8 & 21.4 & 80.0 & 33.8 & 10.1 & 51.0 & 16.9 & 10.9 & 51.9 & 16.9 & 26.6 \\
Llemma-PRM800K-7B\citep{gpt_accelera}           & 36.7 & 71.0 & 48.4 & 39.2 & 47.8 & 43.1 & 33.1 & 25.1 & 28.5 & 35.4 & 31.5 & 33.4 & 38.4 \\
Skywork-PRM-7B\citep{he_2024_16998085}              & 61.8 & 82.9 & 70.8 & 43.8 & 62.2 & 53.6 & 17.9 & 31.9 & 22.9 & 14.0 & 41.9 & 21.0 & 42.1 \\
ReasonEval-7B\citep{xia2024evaluating}               & 26.1 & 95.3 & 41.0 & 35.7 & 77.6 & 48.9 & 27.5 & 55.2 & 36.7 & 27.0 & 60.6 & 37.4 & 41.0 \\
Qwen2.5-Math-7B-PRM800K*\citep{prmlessons}    & 53.1 & 95.3 & 68.2 & 48.0 & 90.1 & 62.6 & 35.7 & 87.3 & 50.7 & 29.8 & 86.1 & 44.3 & 56.5 \\
R-PRM-7B-SFT\citep{she2025rprmreasoningdrivenprocessreward}                & 66.2 & 92.7 & 77.2 & 60.3 & 88.2 & 71.6 & 48.6 & 77.3 & 59.6 & 40.1 & 75.5 & 52.3 & 65.2 \\
R-PRM-7B-DPO\citep{she2025rprmreasoningdrivenprocessreward}                & 72.0 & 91.7 & 80.7 & 71.2 & 83.5 & 76.9 & 60.2 & 67.8 & 63.8 & 55.5 & 65.6 & 60.1 & 70.4 \\
GenPRM-7B\citep{zhao2025genprmscalingtesttimecompute}                   & 61.4 & 92.7 & 73.7 & 70.7 & 86.6 & 77.9 & 68.2 & 75.8 & 71.8 & 68.8 & 79.7 & 73.8 & 74.3 \\

\rowcolor{ourspink}
GPRM-4B-SFT     & 60.5 & 92.1 & 73.1 & 68.4 & 86.0 & 76.2 & 66.0 & 73.1 & 69.4 & 67.3 & 74.0 & 70.5 & 72.3 \\
\rowcolor{ourspink}
GPRM-4B-GRPO       & 61.3 & 92.0 & 73.1 & 70.0 & 86.8 & 77.5 & 68.0 & 75.3 & 71.5 & 70.1 & 80.9 & 75.1 & 74.3 \\
\rowcolor{ourspink}
$\star$ \textbf{GPRM-14B-GRPO} & 62.3 & 93.4 & 74.7 & 72.2 & 88.0 & 79.3 & 70.8 & 77.4 & 73.9 & 72.5 & 78.3 & 75.3 & \textbf{75.8} \\
\bottomrule
\end{tabular}%
}
\end{subtable}

\vspace{0.2cm}
\end{table}

Traditional PRMs judge the $i$-th step as $\hat{y}_i = f_\theta(Q, \mathbf{s}_{1:i})$, a local decision on only the question and current prefix. Yet a step's correctness depends both on prior reasoning (via intermediate conclusions) and on how it is used later (some steps only prove useful in hindsight). We thus formulate step-level evaluation as structured prediction over the full trajectory: given the question $Q$, the historical trajectory $\mathbf{s}_{1:i-1}$, the current step $s_i$, and the subsequent trajectory $\mathbf{s}_{i+1:n}$, the \emph{Global Perspective Process Reward Model} (GPRM) performs structured diagnosis and label prediction for each step using complete contextual information.

\subsection{Global Perspective Process Reward Model}\label{subsec:gprm}

\paragraph{Global Input Design.}
For the $i$-th step, we organize previous steps and their predicted 
judgments into a historical state $\mathcal{H}_i=\{(s_j,\hat{y}_j)\}_{j=1}^{i-1}$, 
where $\mathcal{H}_1 = \emptyset$ for the first step. In practice, 
$\mathcal{H}_i$ is serialized as a text sequence where each entry 
$(s_j, \hat{y}_j)$ is formatted as ``Step $j$: [step text] $\to$ [CORRECT/INCORRECT]'', 
maintaining the temporal order. We denote the subsequent trajectory as 
$\mathcal{F}_i = \mathbf{s}_{i+1:n}$ if $i < n$, and $\mathcal{F}_i = \emptyset$ 
if $i = n$. The unified input of GPRM at step $i$ is then 
$X_i=(Q,\mathcal{H}_i,s_i,\mathcal{F}_i)$.

\begin{table}[tbp] 
\centering
\caption{Performance on \textbf{Agent Error Bench}. \textbf{S}: earliest error step localization. \textbf{S+M}: step + module identification. $\star$ denotes our models.}
\label{tab:agent_error_bench_single}

\footnotesize 
\setlength{\tabcolsep}{2.5pt} 
\renewcommand{\arraystretch}{1.1}

\begin{tabularx}{\linewidth}{@{}l *{8}{>{\centering\arraybackslash}X}@{}}
\toprule
\multirow{2}{*}{\textbf{Method}} 
& \multicolumn{2}{c}{\textbf{ALFWorld}} 
& \multicolumn{2}{c}{\textbf{WebShop}} 
& \multicolumn{2}{c}{\textbf{GAIA}} 
& \multicolumn{2}{c}{\textbf{Avg.}} \\ 
\cmidrule(lr){2-3} \cmidrule(lr){4-5} \cmidrule(lr){6-7} \cmidrule(lr){8-9}
& S & S+M & S & S+M & S & S+M & S & S+M \\
\midrule
\multicolumn{9}{l}{\textit{GPT-4.1-based baselines}} \\
Direct Prompting & 28.0 & 14.0 & 30.0 & 6.0  & 26.0 & 10.0 & 28.0 & 10.0 \\
Brute Force      & 10.0 & 5.0  & 8.0  & 0.0  & 18.0 & 8.0  & 12.0 & 4.3 \\
Binary Search    & 20.0 & 6.0  & 14.0 & 8.0  & 22.0 & 10.0 & 18.7 & 8.0 \\
AgentDebug       & 35.0 & 28.0 & 42.0 & 22.0 & 58.0 & 44.0 & 45.0 & 31.3 \\
\midrule
\rowcolor{gray!15}
$\star$ GPRM-4B  & 38.0 & 30.0 & 44.0 & 24.0 & 60.0 & 46.0 & 47.0 & 33.0 \\
\rowcolor{gray!15}
$\star$ GPRM-14B & \textbf{46.0} & \textbf{37.0} & \textbf{51.0} & \textbf{29.0} & \textbf{67.0} & \textbf{51.0} & \textbf{54.0} & \textbf{39.0} \\
\bottomrule
\end{tabularx}
\end{table}

This global input design enables GPRM to:
\begin{enumerate}[leftmargin=*,topsep=0.5pt,itemsep=0pt,parsep=0pt]
    \item \textbf{Historical Judgments} ($\mathcal{H}_i$): 
          Earlier errors may effect later steps through error spread.
    \item \textbf{Future Trajectory judgments} ($\mathcal{F}_i$): 
          Some steps can only be validated by examining their usage in 
          subsequent reasoning.
    \item \textbf{Maintain Problem Context} ($Q$): Ensure alignment with 
          the original objective.
\end{enumerate}

\paragraph{Four-Dimensional Structured Analysis.}
Given the input $X_i$, the model outputs a structured analysis and the 
final predicted label: $\hat{A}_i,\hat{y}_i = g_\phi(X_i)$, where $g_\phi$ 
is the parameterized process evaluation model. When viewed as a probability 
distribution, we write $P_\phi(\cdot \mid X_i)$; in the reinforcement 
learning context, we use $\pi_\phi$ to denote the policy.

To improve stability and interpretability, we decompose the step analysis 
into four fixed dimensions: $\hat{A}_i = (\hat{a}_i^{\scriptscriptstyle\mathrm{F}}, \hat{a}_i^{\scriptscriptstyle\mathrm{P}}, \hat{a}_i^{\scriptscriptstyle\mathrm{G}}, \hat{a}_i^{\scriptscriptstyle\mathrm{S}})$, 
where the superscripts denote:

\begin{enumerate}[leftmargin=*,topsep=0.5pt,itemsep=0pt,parsep=0pt]
    \item $\mathrm{F}$ (\textbf{Forward Consistency}): checks if
          the current step is consistent with prior $\mathbf{s}_{1:i-1}$.
    \item $\mathrm{P}$ (\textbf{Posterior Verification}): checks 
          if the current step is reasonably used in subsequent 
          $\mathcal{F}_i$.
    \item $\mathrm{G}$ (\textbf{Goal Alignment}): determines if
          the current step still serves the problem $Q$.
    \item $\mathrm{S}$ (\textbf{Self Validity}): verifies if the current step is logically and computationally correct.

\end{enumerate}

The final step label is extracted from the generated analysis via a parsing 
function. Specifically, the model is trained to output judgment 
token at the end of $\hat{A}_i$: $\hat{y}_i = \operatorname{Parse}(\hat{A}_i) = 1$ 
if $\hat{A}_i$ ends with ``[CORRECT]'', and $\hat{y}_i = 0$ if it ends with 
``[INCORRECT]''.

\paragraph{Recursive Evaluation Process.}
After completing the judgment for step $i$, the current prediction is 
written back into the historical state: $\mathcal{H}_{i+1}=\mathcal{H}_i \cup \{(s_i,\hat{y}_i)\}$. 
Therefore, GPRM is no longer a local model that scores each step independently;
instead, it is a global evaluation process that unfolds recursively along the
entire reasoning trajectory.

Because GPRM scores a step only in light of its future trajectory
$\mathcal{F}_i$, evaluating a \emph{candidate} step in online search requires
that the step first be extended into a complete trajectory so that
$\mathcal{F}_i$ is available. In two of our three inference regimes this incurs
no extra generation cost, since $\mathcal{F}_i$ is already contained in the
given trajectory (offline diagnosis) or in the sampled completions
(Best-of-$N$); only the step-wise online search pays for completing
candidates. We quantify this cost in Section~\ref{sec:cost_analysis}.

\subsection{Progressive Training Strategy}\label{subsec:progressive}

We adopt a two-stage progressive training strategy. Both stages use the
complete global input $X_i=(Q,\mathcal{H}_i,s_i,\mathcal{F}_i)$, but differ
in the training objective.

\textbf{Stage I (Supervised Fine-Tuning).} The model learns step-level
structured analysis by supervised imitation of the teacher's four-dimensional
diagnosis.

\textbf{Stage II (Reinforcement Learning).} The model is further optimized
by GRPO with an outcome-based reward, refining its
global judgments under the same input.

Through this two-stage training framework, we have enabled GPRM to possess a four-dimensional step correction capability with a global perspective and historical dependencies.

\subsection{Supervised Fine-Tuning for Step Analysis}\label{subsec:sft}

We construct an SFT dataset based on \textbf{OpenMathInstruct-2}, aligned with the four analysis dimensions by injecting targeted errors and collecting
teacher-verified analyses.

\paragraph{Error Injection.}
Given a seed sample $(Q,\mathbf{s})$ with
$\mathbf{s}=(s_1,s_2,\dots,s_n)$, we define a set of fine-grained error types
$\mathcal{E}$:
\begin{enumerate}[leftmargin=*,topsep=0.5pt,itemsep=0pt,parsep=0pt]
    \item \textbf{Calculation errors}: arithmetic mistakes.
    \item \textbf{Logical errors}: invalid inference steps.
    \item \textbf{Goal deviation}: drifting away from the objective.
    \item \textbf{Inconsistency errors}: contradicting prior conclusions.
\end{enumerate}
For an error type $e\in\mathcal{E}$ and a target position $i$, a teacher model
$\mathcal{T}$ injects an error into step $s_i$ via
$\tilde{s}_i^{(e)}=\mathcal{T}_{\text{inject}}(s_i,e\mid Q,\mathbf{s}_{1:i-1})$,
and we propagate its effect to obtain a perturbed trajectory
$\tilde{\mathbf{s}}^{(e)}=(s_1,\dots,s_{i-1},\tilde{s}_i^{(e)},\tilde{s}_{i+1}^{(e)},\dots,\tilde{s}_n^{(e)})$.

\paragraph{Teacher Annotation and Quality Control.}
Given $(Q,\tilde{\mathbf{s}}^{(e)},i)$, the teacher produces a structured
analysis $A_i=\mathcal{T}_{\text{analyze}}(Q,\tilde{\mathbf{s}}^{(e)},i)$.
If the teacher's judgment is inconsistent with the gold label $y_i^*$, we
prompt for correction and re-analysis to ensure alignment with $y_i^*$.

\paragraph{Dataset Construction.}
We build the SFT dataset
$\mathcal{D}_{\text{SFT}}=\{(X_i^{\text{SFT}},A_i,y_i^*)\}$ over the complete
global input $X_i^{\text{SFT}}=(Q,\mathcal{H}_i,\tilde{s}_i^{(e)},\mathcal{F}_i)$,
where $\mathcal{H}_i$ are the teacher's gold judgments on prior steps and
$\mathcal{F}_i$ is the remainder of the perturbed trajectory, so the model
learns four-dimensional analysis already under historical judgments and the
future view.

\begin{table}[tbp]
\centering
\caption{Performance under Best-of-8 and Greedy Guided Search@8 using Qwen2.5-7B-Instruct as the policy model. $\star$ denotes our models.}
\label{tab:test_time_search}

\footnotesize 
\renewcommand{\arraystretch}{1.0} 
\setlength{\tabcolsep}{2.8pt} 

\begin{tabular}{l ccccccc}
\toprule
\textbf{Setting / Model} & \textbf{AIME24} & \textbf{AMC23} & \textbf{MATH} & \textbf{Olymp. Bench} & \textbf{College Math} & \textbf{Minerva MATH} & \textbf{Avg.} \\
\midrule
\multicolumn{8}{l}{\textit{Reference Baselines}} \\
pass@1 & 11.2 & 47.8 & 73.0 & 38.0 & 38.6 & 37.2 & 41.0 \\
maj@8  & 20.0 & 57.5 & 79.6 & 47.0 & 41.5 & 42.7 & 48.0 \\
pass@8 & 33.3 & 82.5 & 88.8 & 58.5 & 47.5 & 57.7 & 61.4 \\
\midrule
\multicolumn{8}{l}{\textit{Best-of-8}} \\
Math-PSA-7B             & 20.0 & 55.0 & 80.8 & 47.5 & 39.5 & 40.1 & 47.2 \\
RLHFlow-DeepSeek-8B     & 13.3 & 57.5 & 76.2 & 40.0 & 39.0 & 39.7 & 44.3 \\
Skywork-PRM-7B          & 16.7 & 55.0 & 81.2 & 44.0 & 40.5 & 44.5 & 47.0 \\
ReasonEval-7B           & 6.7  & 55.0 & 75.2 & 41.0 & 40.0 & 40.4 & 43.1 \\
Qwen2.5-Math-7B-PRM800K & 13.3 & 57.5 & 80.0 & 44.5 & 43.5 & 43.0 & 47.7 \\
Qwen2.5-Math-PRM-7B     & 16.7 & 55.0 & 82.0 & 48.0 & 43.5 & 43.0 & 48.0 \\
R-PRM-7B-DPO            & 20.0 & 62.5 & 82.2 & 48.0 & 41.0 & 44.1 & 49.6 \\
\rowcolor{gray!25}
$\star$ GPRM-4B         & \textbf{20.0} & \textbf{63.0} & \textbf{82.6} & \textbf{48.5} & 40.5 & \textbf{45.0} & \textbf{50.1} \\
\rowcolor{gray!25}
$\star$ GPRM-14B        & \textbf{20.0} & \textbf{64.2} & \textbf{83.1} & \textbf{50.3} & 42.6 & \textbf{45.8} & \textbf{51.0} \\
\midrule
\multicolumn{8}{l}{\textit{Greedy Guided Search@8}} \\
Math-Shepherd-7B        & 13.3 & 52.5 & 74.6 & 38.5 & 36.5 & 41.2 & 42.8 \\
RLHFlow-PRM-DeepSeek-8B & 13.3 & 52.5 & 74.8 & 39.5 & 37.0 & 40.8 & 43.0 \\
Llemma-PRM800K-7B       & 13.3 & 57.5 & 73.8 & 40.0 & 36.5 & 38.2 & 43.2 \\
Skywork-PRM-7B          & 10.0 & 57.5 & 77.8 & 41.5 & 39.0 & 43.4 & 44.9 \\
ReasonEval-7B           & 3.3  & 55.0 & 73.0 & 37.5 & 35.5 & 37.9 & 40.4 \\
Qwen2.5-Math-7B-PRM800K & 23.3 & 45.0 & 78.2 & 42.0 & 35.5 & 38.6 & 43.8 \\
Qwen2.5-Math-PRM-7B     & 16.7 & 60.0 & 81.0 & 43.5 & 39.0 & 40.4 & 46.8 \\
R-PRM-7B-DPO            & 16.7 & 70.0 & 80.0 & 46.5 & 39.5 & 43.4 & 49.4 \\
\rowcolor{gray!25}
$\star$ GPRM-4B         & \textbf{23.3} & 85.0 & 80.0 & \textbf{48.0} & \textbf{45.0} & 48.8 & \textbf{55.0} \\
\rowcolor{gray!25}
$\star$ GPRM-14B        & \textbf{23.3} & \textbf{87.5} & \textbf{85.0} & 45.0 & 39.5 & \textbf{50.0} & \textbf{55.0} \\
\bottomrule
\end{tabular}
\end{table}

\paragraph{Training Objective.}
We optimize the autoregressive maximum likelihood objective
\[
\mathcal{L}_{\text{SFT}}(\phi)
=
-\sum_{(X_i^{\text{SFT}},A_i)\in\mathcal{D}_{\text{SFT}}}
\log P_\phi(A_i \mid X_i^{\text{SFT}}).
\]

\begin{table}[tbp]
    \centering
    \caption{Ablation study on the impact of historical judgments, future trajectories, and analysis dimensions. Results are reported using GPRM-4B-SFT.}
    \label{tab:ablation}
    
    \footnotesize 
    
    \begin{tabular*}{\linewidth}{@{\extracolsep{\fill}} l cc}
        \toprule
        & \multicolumn{2}{c}{\textbf{Performance Metrics}} \\
        \cmidrule(lr){2-3}
        \textbf{Variant} & \textbf{PRM (Overall)} & \textbf{Proc (Avg. F1)} \\
        \midrule
        
        w/o Hist. \& Future & 66.2\down{6.7} & 57.0\down{15.3} \\
        w/o Hist. Judg.     & 69.8\down{3.1} & 61.1\down{11.2} \\
        w/o Future          & 71.6\down{1.3} & 56.3\down{16.0} \\
        w/o Self Check      & 68.7\down{4.2} & 62.9\down{9.4}  \\
        w/o Goal Align.     & 69.9\down{3.0} & 64.7\down{7.6}  \\
        
        \midrule
        Full GPRM           & \textbf{72.9}  & \textbf{72.3}   \\
        \bottomrule
    \end{tabular*}
    
    \vspace{0.2em}
    \footnotesize{\textit{Note:} ``PRM'' denotes PRMBench, ``Proc'' denotes ProcessBench. ``Hist.'' denotes Historical Judgments.}
\end{table}

\subsection{Reinforcement Learning for Global Evaluation}\label{subsec:grpo}

While SFT already teaches the four-dimensional analysis under the full global input, it is trained only on teacher-verified trajectories. We therefore construct a GRPO dataset based on \textbf{PRM800K}. Since PRM800K truncates each trajectory at the first erroneous step, the future trajectory $\mathcal{F}_i$ is unavailable for posterior verification, so we complete it during data construction. We focus on hard cases with late-occurring errors and refine the model's judgments via an outcome-based reward. We apply GRPO on the same complete global input $X_i^{\text{GRPO}}=(Q,\mathcal{H}_i,s_i,\mathcal{F}_i)$, involving three key stages:

\paragraph{Future Trajectory Completion.}
For an erroneous step $s_i$, we sample candidate continuations $\tilde{\mathcal{F}}_i=(\tilde{s}_{i+1},\ldots,\tilde{s}_{n'})\sim \mathcal{G}_{\text{gen}}(\cdot\mid Q,\mathbf{s}_{1:i-1},s_i)$ from a teacher model, retaining only those yielding incorrect final answers: $\tilde{\mathcal{F}}_i\in\{\mathcal{F}\mid \operatorname{Answer}(\mathcal{F})\neq \operatorname{Answer}^*\}$.

\paragraph{Hard Sample Mining.}
We construct $X_i^{\text{GRPO}}=(Q,\mathcal{H}_i,s_i,\tilde{\mathcal{F}}_i)$, where $\mathcal{H}_i=\{(s_j,\hat{y}_j)\}_{j=1}^{i-1}$ is generated by the SFT model. We then sample predictions $\hat{y}_i^{(k)}\sim P_{\phi_{\text{SFT}}}(y\mid X_i^{\text{GRPO}})$ and retain inputs where at least one prediction disagrees with the gold label, forming the hard dataset $\mathcal{D}_{\text{hard}}=\{X_i^{\text{GRPO}}\mid \exists k,\ \hat{y}_i^{(k)}\neq y_i^*\}$.

\paragraph{GRPO Optimization.}
Let $\pi_\phi$ be the current policy and $\pi_{\phi_{\text{ref}}}$ the reference policy (initialized from SFT). We employ an outcome-based reward $r_i^{(k)}=\mathbb{I}[\operatorname{Parse}(\hat{A}_i^{(k)})=y_i^*]$, where $\operatorname{Parse}(\cdot)$ extracts the final correctness label. The GRPO objective is:
\begin{multline*}
\mathcal{L}_{\text{GRPO}}(\phi)
= -\mathbb{E}_{X_i^{\text{GRPO}}\sim \mathcal{D}_{\text{hard}}}
\Bigg[
\frac{1}{K}\sum_{k=1}^{K}
\min\Big(
\rho_i^{(k)}\operatorname{Adv}_i^{(k)}, \\
\operatorname{clip}(\rho_i^{(k)},1-\epsilon,1+\epsilon)\operatorname{Adv}_i^{(k)}
\Big)
\Bigg],
\end{multline*}
where $\rho_i^{(k)}=\frac{\pi_\phi(\hat{A}_i^{(k)}\mid X_i^{\text{GRPO}})}{\pi_{\phi_{\text{ref}}}(\hat{A}_i^{(k)}\mid X_i^{\text{GRPO}})}$ and $\operatorname{Adv}_i^{(k)}=r_i^{(k)}-\frac{1}{K}\sum_{j=1}^{K}r_i^{(j)}$. This optimization on $\mathcal{D}_{\text{hard}}$ enables robust step-level judgments under complete global context.

\section{Experiments}\label{sec:experiments}

\subsection{Experimental Setup}

\textbf{Models.} We train GPRM based on Qwen3-4B-Instruct and Qwen3-14B (scalability validation). Qwen3-235B-Instruct serves as the teacher model for error injection and annotation.

\textbf{Benchmarks.} We evaluate on process supervision (PRMBench, ProcessBench), cross-task generalization (Agent Error Bench, PPE), and test-time search (Best-of-8, Greedy Guided Search@8). Complete benchmark descriptions are in Appendix~\ref{app:benchmarks}.

\textbf{Training Data and Settings.} We adopt a two-stage training pipeline with standard autoregressive training. Details are in Appendices~\ref{app:data} and~\ref{app:training}.

\subsection{Main Results}

\textbf{Process Supervision.} GPRM-4B achieves an overall score of 73.9 on PRMBench and an average F1 of 74.3 on ProcessBench, outperforming Qwen2.5-Math-PRM-7B (65.5, --), R-PRM-7B-DPO (66.8, 70.4), and proprietary models including GPT-4o (66.8, 61.9) and o1-mini (68.8, --). GPRM-14B further improves to 74.6 and 75.8, respectively.

\textbf{Agent Error Bench.} GPRM-4B achieves 47.0\% on S and 33.0\% on S+M, surpassing AgentDebug (45.0\%, 31.3\%). GPRM-14B further improves to 54.0\% and 39.0\%, demonstrating effective transfer to agent scenarios. We use Agent Error Bench to demonstrate cross-task generalization of the four-dimensional global framework; evaluating the end-to-end task success rate when GPRM is integrated as a verifier in interactive environments such as ALFWorld and GAIA is left to future work. Beyond agent and mathematical reasoning, GPRM also shows strong generalization to reasoning tasks (see Appendix~\ref{sec:ppe_generalization} for PPE benchmark results).

\subsection{Downstream Test-Time Search Results}

\textbf{Test-Time Search.} Table~\ref{tab:test_time_search} shows results under Best-of-8 and Greedy Guided Search@8 using Qwen2.5-7B-Instruct. Under Best-of-8, GPRM-4B achieves an average of \textbf{50.1}, outperforming R-PRM-7B-DPO (\textbf{49.6}). GPRM-14B further improves to \textbf{51.0}. The gain is moderate as Best-of-$N$ is limited by the quality of the candidate pool. Under Greedy Guided Search@8, both GPRM-4B and GPRM-14B achieve an average of \textbf{55.0}, significantly outperforming R-PRM-7B-DPO (\textbf{49.4}).

\subsection{Inference Cost Analysis}\label{sec:cost_analysis}

We analyze the inference cost introduced by the future-step view $\mathcal{F}_i$. We compare the identical GPRM-4B-SFT with and without $\mathcal{F}_i$ (same Qwen3-4B base, same training, same 8$\times$H20-96\,GB) over three regimes (Table~\ref{tab:cost}). The key distinction is in online search: without $\mathcal{F}_i$ a standard greedy pass scores 8 single-step candidates directly, whereas with $\mathcal{F}_i$ each of the 8 candidates must first be completed into a full trajectory before GPRM scores it---it is this completion, not the future text, that raises cost.

\begin{table}[t]
\centering
\caption{Inference cost with/without future trajectory $\mathcal{F}_i$. Measured on GPRM-4B-SFT (8$\times$H20, 96GB). Tokens: input+output per problem. Latency: end-to-end per problem. Score: mean of PRMBench/ProcessBench (diagnosis) or avg. of six TTS benchmarks. ``w/o $\mathcal{F}_i$'' is control; differences arise solely from whether $\mathcal{F}_i$ is used.}
\label{tab:cost}
\small
\setlength{\tabcolsep}{2pt}
\renewcommand{\arraystretch}{1.0}
\begin{tabularx}{\linewidth}{@{}l >{\centering\arraybackslash}X *{4}{>{\raggedleft\arraybackslash}X}@{}}
\toprule
\textbf{Regime} & \textbf{$\mathcal{F}_i$} & \textbf{Tok/q} & \textbf{Lat(s)} & \textbf{Tok+ (\%)} & \textbf{Score} \\
\midrule
\multicolumn{6}{l}{\textit{Single-problem diag. (offline)}} \\
\quad GPRM-4B (w/o $\mathcal{F}_i$) & \ding{55} & 17555 & 12.5 & --    & 61.6 \\
\quad GPRM-4B (ours)              & \ding{51} & 19253 & 17.8 & +9.6  & 72.6 \\
\midrule
\multicolumn{6}{l}{\textit{Best-of-8 (offline TTS)}} \\
\quad GPRM-4B (w/o $\mathcal{F}_i$) & \ding{55} & 140442 & 15.5 & --     & 47.8 \\
\quad GPRM-4B (ours)              & \ding{51} & 162755 & 18.7 & +15.9  & 50.0 \\
\midrule
\multicolumn{6}{l}{\textit{Greedy Guided Search@8 (online TTS)}} \\
\quad GPRM-4B (w/o $\mathcal{F}_i$) & \ding{55} & 152327 & 17.7 & --     & 48.6 \\
\quad GPRM-4B (ours)              & \ding{51} & 261532 & 25.2 & +71.7  & 55.0 \\
\bottomrule
\end{tabularx}
\end{table}

The cost of $\mathcal{F}_i$ is \emph{regime-localized}. It is nearly free offline (+9.6\% tokens, +11.1 pts) and in Best-of-8 (+15.9\% tokens, +2.2 pts), since the future is already part of the given or sampled trajectory. Only online greedy search is costly (+71.7\% tokens, +42.4\% latency) because of candidate completion---but it buys +6.4 pts, consistent with the 16.0-pt drop when $\mathcal{F}_i$ is removed (Table~\ref{tab:ablation}). The overhead is thus the price of the global view, confined to online search. 

\subsection{Ablation Studies}

Table~\ref{tab:ablation} presents ablation results on PRMBench and ProcessBench using GPRM-4B-SFT to isolate component effects. As Table~\ref{tab:process_supervision} shows, the SFT$\to$GRPO transition consistently improves performance, suggesting these findings remain valid post-GRPO. We treat the two-stage pipeline as a design necessity rather than an ablation variable: direct GRPO application fails to converge due to the complexity of simultaneously learning four-dimensional diagnostics and structured formatting under high-dimensional global input $X_i=(Q,\mathcal{H}_i,s_i,\mathcal{F}_i)$. SFT establishes the foundational analysis skills and format, enabling GRPO to effectively refine judgments on hard cases.

\textbf{Impact of Global Context.} Removing both historical judgments and future trajectories causes significant drops of 6.7 (PRMBench) and 15.3 (ProcessBench) points. Isolating these components reveals that removing only historical judgments leads to drops of 3.1 and 11.2 points, while removing only future trajectories causes drops of 1.3 and 16.0 points, highlighting the critical role of future trajectories for early error localization in ProcessBench.

\textbf{Impact of Analysis Dimensions.} Removing Self Check and Goal Alignment results in performance drops of 4.2/9.4 and 3.0/7.6 points, respectively. These results confirm the contribution of all four dimensions. Independent expert review further validated the F/P/G/S axes as orthogonal and comprehensive in covering common practitioner-identified failure modes.

\section{Conclusion}

We propose GPRM, a global perspective process reward model that addresses two critical limitations of existing PRMs: the lack of historical judgment inheritance and the absence of a global perspective. By jointly conditioning on historical judgments, the current step, and future trajectories, GPRM enables robust step-level evaluation through a structured four-dimensional diagnostic framework. Our progressive training pipeline combines supervised fine-tuning with hard-example-based GRPO to enhance both foundational analysis capabilities and decision robustness in challenging scenarios. Extensive experiments demonstrate that GPRM achieves state-of-the-art performance on process supervision benchmarks, generalizes effectively to agent error localization and general reward modeling tasks, and significantly improves test-time search guidance.

\FloatBarrier
\bibliographystyle{abbrvnat}
\bibliography{references}

\begin{thebibliography}{32}
\providecommand{\natexlab}[1]{#1}
\providecommand{\url}[1]{\texttt{#1}}
\expandafter\ifx\csname urlstyle\endcsname\relax
  \providecommand{\doi}[1]{doi: #1}\else
  \providecommand{\doi}{doi: \begingroup \urlstyle{rm}\Url}\fi

\bibitem[Chen et~al.(2026)Chen, Li, Wang, Jin, Qian, Wang, Wang, Zhang, Zhang,
  Zhang, Tong, and Ji]{chen2026rmr1rewardmodelingreasoning}
X.~Chen, G.~Li, Z.~Wang, B.~Jin, C.~Qian, Y.~Wang, H.~Wang, Y.~Zhang, D.~Zhang,
  T.~Zhang, H.~Tong, and H.~Ji.
\newblock Rm-r1: Reward modeling as reasoning, 2026.
\newblock URL \url{https://arxiv.org/abs/2505.02387}.

\bibitem[Frick et~al.(2025)Frick, Li, Chen, Chiang, Angelopoulos, Jiao, Zhu,
  Gonzalez, and Stoica]{ppe_citation}
E.~Frick, T.~Li, C.~Chen, W.-L. Chiang, A.~N. Angelopoulos, J.~Jiao, B.~Zhu,
  J.~E. Gonzalez, and I.~Stoica.
\newblock How to evaluate reward models for {RLHF}.
\newblock In \emph{The Thirteenth International Conference on Learning
  Representations}, 2025.
\newblock URL \url{https://openreview.net/forum?id=cbttLtO94Q}.

\bibitem[Grattafiori et~al.(2024)Grattafiori, Dubey, Jauhri, Pandey, Kadian,
  Al-Dahle, Letman, Mathur, Schelten, Vaughan, Yang, Fan, Goyal, Hartshorn,
  Yang, Mitra, Sravankumar, Korenev, Hinsvark, Rao, Zhang, Rodriguez,
  Gregerson, Spataru, Roziere, Biron, Tang, Chern, Caucheteux, Nayak, Bi,
  Marra, McConnell, Keller, Touret, Wu, Wong, Ferrer, Nikolaidis, Allonsius,
  Song, Pintz, Livshits, Wyatt, Esiobu, Choudhary, Mahajan, Garcia-Olano,
  Perino, Hupkes, Lakomkin, AlBadawy, Lobanova, Dinan, Smith, Radenovic,
  Guzm{\'a}n, Zhang, Synnaeve, Lee, Anderson, Thattai, Nail, Mialon, Pang,
  Cucurell, Nguyen, Korevaar, Xu, Touvron, Zarov, Ibarra, Kloumann, Misra,
  Evtimov, Zhang, Copet, Lee, Geffert, Vranes, Park, Mahadeokar, Shah, van~der
  Linde, Billock, Hong, Lee, Fu, Chi, Huang, Liu, Wang, Yu, Bitton, Spisak,
  Park, Rocca, Johnstun, Saxe, Jia, Alwala, Prasad, Upasani, Plawiak, Li,
  Heafield, Stone, El-Arini, Iyer, Malik, Chiu, Bhalla, Lakhotia,
  Rantala-Yeary, van~der Maaten, Chen, Tan, Jenkins, Martin, Madaan, Malo,
  Blecher, Landzaat, de~Oliveira, Muzzi, Pasupuleti, Singh, Paluri, Kardas,
  Tsimpoukelli, Oldham, Rita, Pavlova, Kambadur, Lewis, Si, Singh, Hassan,
  Goyal, Torabi, Bashlykov, Bogoychev, Chatterji, Zhang, Duchenne,
  {\c{C}}elebi, Alrassy, Zhang, Li, Vasic, Weng, Bhargava, Dubal, Krishnan,
  Koura, Xu, He, Dong, Srinivasan, Ganapathy, Calderer, Cabral, Stojnic,
  Raileanu, Maheswari, Girdhar, Patel, Sauvestre, Polidoro, Sumbaly, Taylor,
  Silva, Hou, Wang, Hosseini, Chennabasappa, Singh, Bell, Kim, Edunov, Nie,
  Narang, Raparthy, Shen, Wan, Bhosale, Zhang, Vandenhende, Batra, Whitman,
  Sootla, Collot, Gururangan, Borodinsky, Herman, Fowler, Sheasha, Georgiou,
  Scialom, Speckbacher, Mihaylov, Xiao, Karn, Goswami, Gupta, Ramanathan,
  Kerkez, Gonguet, Do, Vogeti, Albiero, Petrovic, Chu, Xiong, Fu, Meers,
  Martinet, Wang, Wang, Tan, Xia, Xie, Jia, Wang, Goldschlag, Gaur, Babaei,
  Wen, Song, Zhang, Li, Mao, Coudert, Yan, Chen, Papakipos, Singh, Srivastava,
  Jain, Kelsey, Shajnfeld, Gangidi, Victoria, Goldstand, Menon, Sharma,
  Boesenberg, Baevski, Feinstein, Kallet, Sangani, Teo, Yunus, Lupu, Alvarado,
  Caples, Gu, Ho, Poulton, Ryan, Ramchandani, Dong, Franco, Goyal, Saraf,
  Chowdhury, Gabriel, Bharambe, Eisenman, Yazdan, James, Maurer, Leonhardi,
  Huang, Loyd, Paola, Paranjape, Liu, Wu, Ni, Hancock, Wasti, Spence,
  Stojkovic, Gamido, Montalvo, Parker, Burton, Mejia, Liu, Wang, Kim, Zhou, Hu,
  Chu, Cai, Tindal, Feichtenhofer, Gao, Civin, Beaty, Kreymer, Li, Adkins, Xu,
  Testuggine, David, Parikh, Liskovich, Foss, Wang, Le, Holland, Dowling,
  Jamil, Montgomery, Presani, Hahn, Wood, Le, Brinkman, Arcaute, Dunbar,
  Smothers, Sun, Kreuk, Tian, Kokkinos, Ozgenel, Caggioni, Kanayet, Seide,
  Florez, Schwarz, Badeer, Swee, Halpern, Herman, Sizov, Guangyi, Zhang,
  Lakshminarayanan, Inan, Shojanazeri, Zou, Wang, Zha, Habeeb, Rudolph, Suk,
  Aspegren, Goldman, Zhan, Damlaj, Molybog, Tufanov, Leontiadis, Veliche, Gat,
  Weissman, Geboski, Kohli, Lam, Asher, Gaya, Marcus, Tang, Chan, Zhen,
  Reizenstein, Teboul, Zhong, Jin, Yang, Cummings, Carvill, Shepard, McPhie,
  Torres, Ginsburg, Wang, Wu, U, Saxena, Khandelwal, Zand, Matosich,
  Veeraraghavan, Michelena, Li, Jagadeesh, Huang, Chawla, Huang, Chen, Garg, A,
  Silva, Bell, Zhang, Guo, Yu, Moshkovich, Wehrstedt, Khabsa, Avalani, Bhatt,
  Mankus, Hasson, Lennie, Reso, Groshev, Naumov, Lathi, Keneally, Liu, Seltzer,
  Valko, Restrepo, Patel, Vyatskov, Samvelyan, Clark, Macey, Wang, Hermoso,
  Metanat, Rastegari, Bansal, Santhanam, Parks, White, Bawa, Singhal, Egebo,
  Usunier, Mehta, Laptev, Dong, Cheng, Chernoguz, Hart, Salpekar, Kalinli,
  Kent, Parekh, Saab, Balaji, Rittner, Bontrager, Roux, Dollar, Zvyagina,
  Ratanchandani, Yuvraj, Liang, Alao, Rodriguez, Ayub, Murthy, Nayani, Mitra,
  Parthasarathy, Li, Hogan, Battey, Wang, Howes, Rinott, Mehta, Siby, Bondu,
  Datta, Chugh, Hunt, Dhillon, Sidorov, Pan, Mahajan, Verma, Yamamoto,
  Ramaswamy, Lindsay, Lindsay, Feng, Lin, Zha, Patil, Shankar, Zhang, Zhang,
  Wang, Agarwal, Sajuyigbe, Chintala, Max, Chen, Kehoe, Satterfield,
  Govindaprasad, Gupta, Deng, Cho, Virk, Subramanian, Choudhury, Goldman,
  Remez, Glaser, Best, Koehler, Robinson, Li, Zhang, Matthews, Chou, Shaked,
  Vontimitta, Ajayi, Montanez, Mohan, Kumar, Mangla, Ionescu, Poenaru,
  Mihailescu, Ivanov, Li, Wang, Jiang, Bouaziz, Constable, Tang, Wu, Wang, Wu,
  Gao, Kleinman, Chen, Hu, Jia, Qi, Li, Zhang, Zhang, Adi, Nam, Yu, Wang, Zhao,
  Hao, Qian, Li, He, Rait, DeVito, Rosnbrick, Wen, Yang, Zhao, and
  Ma]{grattafiori2024llama3herdmodels}
A.~Grattafiori, A.~Dubey, A.~Jauhri, A.~Pandey, A.~Kadian, A.~Al-Dahle,
  A.~Letman, A.~Mathur, A.~Schelten, A.~Vaughan, A.~Yang, A.~Fan, A.~Goyal,
  A.~Hartshorn, A.~Yang, A.~Mitra, A.~Sravankumar, A.~Korenev, A.~Hinsvark,
  A.~Rao, A.~Zhang, A.~Rodriguez, A.~Gregerson, A.~Spataru, B.~Roziere,
  B.~Biron, B.~Tang, B.~Chern, C.~Caucheteux, C.~Nayak, C.~Bi, C.~Marra,
  C.~McConnell, C.~Keller, C.~Touret, C.~Wu, C.~Wong, C.~C. Ferrer,
  C.~Nikolaidis, D.~Allonsius, D.~Song, D.~Pintz, D.~Livshits, D.~Wyatt,
  D.~Esiobu, D.~Choudhary, D.~Mahajan, D.~Garcia-Olano, D.~Perino, D.~Hupkes,
  E.~Lakomkin, E.~AlBadawy, E.~Lobanova, E.~Dinan, E.~M. Smith, F.~Radenovic,
  F.~Guzm{\'a}n, F.~Zhang, G.~Synnaeve, G.~Lee, G.~L. Anderson, G.~Thattai,
  G.~Nail, G.~Mialon, G.~Pang, G.~Cucurell, H.~Nguyen, H.~Korevaar, H.~Xu,
  H.~Touvron, I.~Zarov, I.~A. Ibarra, I.~Kloumann, I.~Misra, I.~Evtimov,
  J.~Zhang, J.~Copet, J.~Lee, J.~Geffert, J.~Vranes, J.~Park, J.~Mahadeokar,
  J.~Shah, J.~van~der Linde, J.~Billock, J.~Hong, J.~Lee, J.~Fu, J.~Chi,
  J.~Huang, J.~Liu, J.~Wang, J.~Yu, J.~Bitton, J.~Spisak, J.~Park, J.~Rocca,
  J.~Johnstun, J.~Saxe, J.~Jia, K.~V. Alwala, K.~Prasad, K.~Upasani,
  K.~Plawiak, K.~Li, K.~Heafield, K.~Stone, K.~El-Arini, K.~Iyer, K.~Malik,
  K.~Chiu, K.~Bhalla, K.~Lakhotia, L.~Rantala-Yeary, L.~van~der Maaten,
  L.~Chen, L.~Tan, L.~Jenkins, L.~Martin, L.~Madaan, L.~Malo, L.~Blecher,
  L.~Landzaat, L.~de~Oliveira, M.~Muzzi, M.~Pasupuleti, M.~Singh, M.~Paluri,
  M.~Kardas, M.~Tsimpoukelli, M.~Oldham, M.~Rita, M.~Pavlova, M.~Kambadur,
  M.~Lewis, M.~Si, M.~K. Singh, M.~Hassan, N.~Goyal, N.~Torabi, N.~Bashlykov,
  N.~Bogoychev, N.~Chatterji, N.~Zhang, O.~Duchenne, O.~{\c{C}}elebi,
  P.~Alrassy, P.~Zhang, P.~Li, P.~Vasic, P.~Weng, P.~Bhargava, P.~Dubal,
  P.~Krishnan, P.~S. Koura, P.~Xu, Q.~He, Q.~Dong, R.~Srinivasan, R.~Ganapathy,
  R.~Calderer, R.~S. Cabral, R.~Stojnic, R.~Raileanu, R.~Maheswari, R.~Girdhar,
  R.~Patel, R.~Sauvestre, R.~Polidoro, R.~Sumbaly, R.~Taylor, R.~Silva, R.~Hou,
  R.~Wang, S.~Hosseini, S.~Chennabasappa, S.~Singh, S.~Bell, S.~S. Kim,
  S.~Edunov, S.~Nie, S.~Narang, S.~Raparthy, S.~Shen, S.~Wan, S.~Bhosale,
  S.~Zhang, S.~Vandenhende, S.~Batra, S.~Whitman, S.~Sootla, S.~Collot,
  S.~Gururangan, S.~Borodinsky, T.~Herman, T.~Fowler, T.~Sheasha, T.~Georgiou,
  T.~Scialom, T.~Speckbacher, T.~Mihaylov, T.~Xiao, U.~Karn, V.~Goswami,
  V.~Gupta, V.~Ramanathan, V.~Kerkez, V.~Gonguet, V.~Do, V.~Vogeti, V.~Albiero,
  V.~Petrovic, W.~Chu, W.~Xiong, W.~Fu, W.~Meers, X.~Martinet, X.~Wang,
  X.~Wang, X.~E. Tan, X.~Xia, X.~Xie, X.~Jia, X.~Wang, Y.~Goldschlag, Y.~Gaur,
  Y.~Babaei, Y.~Wen, Y.~Song, Y.~Zhang, Y.~Li, Y.~Mao, Z.~D. Coudert, Z.~Yan,
  Z.~Chen, Z.~Papakipos, A.~Singh, A.~Srivastava, A.~Jain, A.~Kelsey,
  A.~Shajnfeld, A.~Gangidi, A.~Victoria, A.~Goldstand, A.~Menon, A.~Sharma,
  A.~Boesenberg, A.~Baevski, A.~Feinstein, A.~Kallet, A.~Sangani, A.~Teo,
  A.~Yunus, A.~Lupu, A.~Alvarado, A.~Caples, A.~Gu, A.~Ho, A.~Poulton, A.~Ryan,
  A.~Ramchandani, A.~Dong, A.~Franco, A.~Goyal, A.~Saraf, A.~Chowdhury,
  A.~Gabriel, A.~Bharambe, A.~Eisenman, A.~Yazdan, B.~James, B.~Maurer,
  B.~Leonhardi, B.~Huang, B.~Loyd, B.~D. Paola, B.~Paranjape, B.~Liu, B.~Wu,
  B.~Ni, B.~Hancock, B.~Wasti, B.~Spence, B.~Stojkovic, B.~Gamido, B.~Montalvo,
  C.~Parker, C.~Burton, C.~Mejia, C.~Liu, C.~Wang, C.~Kim, C.~Zhou, C.~Hu,
  C.-H. Chu, C.~Cai, C.~Tindal, C.~Feichtenhofer, C.~Gao, D.~Civin, D.~Beaty,
  D.~Kreymer, D.~Li, D.~Adkins, D.~Xu, D.~Testuggine, D.~David, D.~Parikh,
  D.~Liskovich, D.~Foss, D.~Wang, D.~Le, D.~Holland, E.~Dowling, E.~Jamil,
  E.~Montgomery, E.~Presani, E.~Hahn, E.~Wood, E.-T. Le, E.~Brinkman,
  E.~Arcaute, E.~Dunbar, E.~Smothers, F.~Sun, F.~Kreuk, F.~Tian, F.~Kokkinos,
  F.~Ozgenel, F.~Caggioni, F.~Kanayet, F.~Seide, G.~M. Florez, G.~Schwarz,
  G.~Badeer, G.~Swee, G.~Halpern, G.~Herman, G.~Sizov, Guangyi, Zhang,
  G.~Lakshminarayanan, H.~Inan, H.~Shojanazeri, H.~Zou, H.~Wang, H.~Zha,
  H.~Habeeb, H.~Rudolph, H.~Suk, H.~Aspegren, H.~Goldman, H.~Zhan, I.~Damlaj,
  I.~Molybog, I.~Tufanov, I.~Leontiadis, I.-E. Veliche, I.~Gat, J.~Weissman,
  J.~Geboski, J.~Kohli, J.~Lam, J.~Asher, J.-B. Gaya, J.~Marcus, J.~Tang,
  J.~Chan, J.~Zhen, J.~Reizenstein, J.~Teboul, J.~Zhong, J.~Jin, J.~Yang,
  J.~Cummings, J.~Carvill, J.~Shepard, J.~McPhie, J.~Torres, J.~Ginsburg,
  J.~Wang, K.~Wu, K.~H. U, K.~Saxena, K.~Khandelwal, K.~Zand, K.~Matosich,
  K.~Veeraraghavan, K.~Michelena, K.~Li, K.~Jagadeesh, K.~Huang, K.~Chawla,
  K.~Huang, L.~Chen, L.~Garg, L.~A, L.~Silva, L.~Bell, L.~Zhang, L.~Guo, L.~Yu,
  L.~Moshkovich, L.~Wehrstedt, M.~Khabsa, M.~Avalani, M.~Bhatt, M.~Mankus,
  M.~Hasson, M.~Lennie, M.~Reso, M.~Groshev, M.~Naumov, M.~Lathi, M.~Keneally,
  M.~Liu, M.~L. Seltzer, M.~Valko, M.~Restrepo, M.~Patel, M.~Vyatskov,
  M.~Samvelyan, M.~Clark, M.~Macey, M.~Wang, M.~J. Hermoso, M.~Metanat,
  M.~Rastegari, M.~Bansal, N.~Santhanam, N.~Parks, N.~White, N.~Bawa,
  N.~Singhal, N.~Egebo, N.~Usunier, N.~Mehta, N.~P. Laptev, N.~Dong, N.~Cheng,
  O.~Chernoguz, O.~Hart, O.~Salpekar, O.~Kalinli, P.~Kent, P.~Parekh, P.~Saab,
  P.~Balaji, P.~Rittner, P.~Bontrager, P.~Roux, P.~Dollar, P.~Zvyagina,
  P.~Ratanchandani, P.~Yuvraj, Q.~Liang, R.~Alao, R.~Rodriguez, R.~Ayub,
  R.~Murthy, R.~Nayani, R.~Mitra, R.~Parthasarathy, R.~Li, R.~Hogan, R.~Battey,
  R.~Wang, R.~Howes, R.~Rinott, S.~Mehta, S.~Siby, S.~J. Bondu, S.~Datta,
  S.~Chugh, S.~Hunt, S.~Dhillon, S.~Sidorov, S.~Pan, S.~Mahajan, S.~Verma,
  S.~Yamamoto, S.~Ramaswamy, S.~Lindsay, S.~Lindsay, S.~Feng, S.~Lin, S.~C.
  Zha, S.~Patil, S.~Shankar, S.~Zhang, S.~Zhang, S.~Wang, S.~Agarwal,
  S.~Sajuyigbe, S.~Chintala, S.~Max, S.~Chen, S.~Kehoe, S.~Satterfield,
  S.~Govindaprasad, S.~Gupta, S.~Deng, S.~Cho, S.~Virk, S.~Subramanian,
  S.~Choudhury, S.~Goldman, T.~Remez, T.~Glaser, T.~Best, T.~Koehler,
  T.~Robinson, T.~Li, T.~Zhang, T.~Matthews, T.~Chou, T.~Shaked, V.~Vontimitta,
  V.~Ajayi, V.~Montanez, V.~Mohan, V.~S. Kumar, V.~Mangla, V.~Ionescu,
  V.~Poenaru, V.~T. Mihailescu, V.~Ivanov, W.~Li, W.~Wang, W.~Jiang,
  W.~Bouaziz, W.~Constable, X.~Tang, X.~Wu, X.~Wang, X.~Wu, X.~Gao,
  Y.~Kleinman, Y.~Chen, Y.~Hu, Y.~Jia, Y.~Qi, Y.~Li, Y.~Zhang, Y.~Zhang,
  Y.~Adi, Y.~Nam, Yu, Wang, Y.~Zhao, Y.~Hao, Y.~Qian, Y.~Li, Y.~He, Z.~Rait,
  Z.~DeVito, Z.~Rosnbrick, Z.~Wen, Z.~Yang, Z.~Zhao, and Z.~Ma.
\newblock The llama 3 herd of models, 2024.
\newblock URL \url{https://arxiv.org/abs/2407.21783}.

\bibitem[Guo et~al.(2025)Guo, Chi, Dong, Dong, Wu, Huang, and
  Wei]{guo2025rewardreasoningmodel}
J.~Guo, Z.~Chi, L.~Dong, Q.~Dong, X.~Wu, S.~Huang, and F.~Wei.
\newblock Reward reasoning model, 2025.
\newblock URL \url{https://arxiv.org/abs/2505.14674}.

\bibitem[He et~al.(2024)He, Wei, Yan, Liu, Wang, Gan, Tu, Liu, Zeng, Wang,
  Wang, Li, Zhang, Xu, An, Liu, and Zhou]{he_2024_16998085}
J.~He, T.~Wei, R.~Yan, J.~Liu, C.~Wang, Y.~Gan, S.~Tu, C.~Y. Liu, L.~Zeng,
  X.~Wang, B.~Wang, Y.~Li, F.~Zhang, J.~Xu, B.~An, Y.~Liu, and Y.~Zhou.
\newblock Skywork-o1 open series, Nov. 2024.
\newblock URL \url{https://doi.org/10.5281/zenodo.16998085}.

\bibitem[Khalifa et~al.(2025)Khalifa, Agarwal, Logeswaran, Kim, Peng, Lee, Lee,
  and Wang]{khalifa2025processrewardmodelsthink}
M.~Khalifa, R.~Agarwal, L.~Logeswaran, J.~Kim, H.~Peng, M.~Lee, H.~Lee, and
  L.~Wang.
\newblock Process reward models that think, 2025.
\newblock URL \url{https://arxiv.org/abs/2504.16828}.

\bibitem[Li and Li(2025)]{li2025processrewardmodelqvalue}
W.~Li and Y.~Li.
\newblock Process reward model with q-value rankings, 2025.
\newblock URL \url{https://arxiv.org/abs/2410.11287}.

\bibitem[Lightman et~al.(2023)Lightman, Kosaraju, Burda, Edwards, Baker, Lee,
  Leike, Schulman, Sutskever, and Cobbe]{lightman2023letsverifystepstep}
H.~Lightman, V.~Kosaraju, Y.~Burda, H.~Edwards, B.~Baker, T.~Lee, J.~Leike,
  J.~Schulman, I.~Sutskever, and K.~Cobbe.
\newblock Let's verify step by step, 2023.
\newblock URL \url{https://arxiv.org/abs/2305.20050}.

\bibitem[Liu et~al.(2025)Liu, Gao, Zhao, Zhang, Li, Qi, Ouyang, and
  Zhou]{liu20251bllmsurpass405b}
R.~Liu, J.~Gao, J.~Zhao, K.~Zhang, X.~Li, B.~Qi, W.~Ouyang, and B.~Zhou.
\newblock Can 1b llm surpass 405b llm? rethinking compute-optimal test-time
  scaling, 2025.
\newblock URL \url{https://arxiv.org/abs/2502.06703}.

\bibitem[Luo et~al.(2024)Luo, Liu, Liu, Phatale, Guo, Lara, Li, Shu, Zhu, Meng,
  Sun, and Rastogi]{luo2024improvemathematicalreasoninglanguage}
L.~Luo, Y.~Liu, R.~Liu, S.~Phatale, M.~Guo, H.~Lara, Y.~Li, L.~Shu, Y.~Zhu,
  L.~Meng, J.~Sun, and A.~Rastogi.
\newblock Improve mathematical reasoning in language models by automated
  process supervision, 2024.
\newblock URL \url{https://arxiv.org/abs/2406.06592}.

\bibitem[Park et~al.(2024)Park, Liu, Gong, and
  Choi]{park2024ensemblinglargelanguagemodels}
S.~Park, X.~Liu, Y.~Gong, and E.~Choi.
\newblock Ensembling large language models with process reward-guided tree
  search for better complex reasoning, 2024.
\newblock URL \url{https://arxiv.org/abs/2412.15797}.

\bibitem[Qwen et~al.(2025)Qwen, :, Yang, Yang, Zhang, Hui, Zheng, Yu, Li, Liu,
  Huang, Wei, Lin, Yang, Tu, Zhang, Yang, Yang, Zhou, Lin, Dang, Lu, Bao, Yang,
  Yu, Li, Xue, Zhang, Zhu, Men, Lin, Li, Tang, Xia, Ren, Ren, Fan, Su, Zhang,
  Wan, Liu, Cui, Zhang, and Qiu]{qwen2025qwen25technicalreport}
Qwen, :, A.~Yang, B.~Yang, B.~Zhang, B.~Hui, B.~Zheng, B.~Yu, C.~Li, D.~Liu,
  F.~Huang, H.~Wei, H.~Lin, J.~Yang, J.~Tu, J.~Zhang, J.~Yang, J.~Yang,
  J.~Zhou, J.~Lin, K.~Dang, K.~Lu, K.~Bao, K.~Yang, L.~Yu, M.~Li, M.~Xue,
  P.~Zhang, Q.~Zhu, R.~Men, R.~Lin, T.~Li, T.~Tang, T.~Xia, X.~Ren, X.~Ren,
  Y.~Fan, Y.~Su, Y.~Zhang, Y.~Wan, Y.~Liu, Z.~Cui, Z.~Zhang, and Z.~Qiu.
\newblock Qwen2.5 technical report, 2025.
\newblock URL \url{https://arxiv.org/abs/2412.15115}.

\bibitem[Setlur et~al.(2024)Setlur, Nagpal, Fisch, Geng, Eisenstein, Agarwal,
  Agarwal, Berant, and Kumar]{setlur2024rewardingprogressscalingautomated}
A.~Setlur, C.~Nagpal, A.~Fisch, X.~Geng, J.~Eisenstein, R.~Agarwal, A.~Agarwal,
  J.~Berant, and A.~Kumar.
\newblock Rewarding progress: Scaling automated process verifiers for llm
  reasoning, 2024.
\newblock URL \url{https://arxiv.org/abs/2410.08146}.

\bibitem[Shao et~al.(2024)Shao, Wang, Zhu, Xu, Song, Bi, Zhang, Zhang, Li, Wu,
  and Guo]{shao2024deepseekmathpushinglimitsmathematical}
Z.~Shao, P.~Wang, Q.~Zhu, R.~Xu, J.~Song, X.~Bi, H.~Zhang, M.~Zhang, Y.~K. Li,
  Y.~Wu, and D.~Guo.
\newblock Deepseekmath: Pushing the limits of mathematical reasoning in open
  language models, 2024.
\newblock URL \url{https://arxiv.org/abs/2402.03300}.

\bibitem[She et~al.(2025)She, Liu, Liu, Chen, Huang, and
  Huang]{she2025rprmreasoningdrivenprocessreward}
S.~She, J.~Liu, Y.~Liu, J.~Chen, X.~Huang, and S.~Huang.
\newblock R-prm: Reasoning-driven process reward modeling, 2025.
\newblock URL \url{https://arxiv.org/abs/2503.21295}.

\bibitem[Song et~al.(2025)Song, Su, Qu, Zhou, and
  Cheng]{song2025prmbenchfinegrainedchallengingbenchmark}
M.~Song, Z.~Su, X.~Qu, J.~Zhou, and Y.~Cheng.
\newblock Prmbench: A fine-grained and challenging benchmark for process-level
  reward models, 2025.
\newblock URL \url{https://arxiv.org/abs/2501.03124}.

\bibitem[Sun(2024)]{gpt_accelera}
Z.~Sun.
\newblock Gpt-accelera: Simple and efficient pytorch-native transformer
  training and inference (batched).
\newblock \url{https://github.com/Edward-Sun/gpt-accelera}, 2024.

\bibitem[Team(2024)]{qwq-32b-preview}
Q.~Team.
\newblock Qwq: Reflect deeply on the boundaries of the unknown, November 2024.
\newblock URL \url{https://qwenlm.github.io/blog/qwq-32b-preview/}.

\bibitem[Wang et~al.(2024)Wang, Li, Shao, Xu, Dai, Li, Chen, Wu, and
  Sui]{wang2024mathshepherdverifyreinforcellms}
P.~Wang, L.~Li, Z.~Shao, R.~X. Xu, D.~Dai, Y.~Li, D.~Chen, Y.~Wu, and Z.~Sui.
\newblock Math-shepherd: Verify and reinforce llms step-by-step without human
  annotations, 2024.
\newblock URL \url{https://arxiv.org/abs/2312.08935}.

\bibitem[Xia et~al.(2024)Xia, Li, Liu, Wu, and Liu]{xia2024evaluating}
S.~Xia, X.~Li, Y.~Liu, T.~Wu, and P.~Liu.
\newblock Evaluating mathematical reasoning beyond accuracy.
\newblock \emph{arXiv preprint arXiv:2404.05692}, 2024.

\bibitem[Xiong et~al.(2024)Xiong, Zhang, Jiang, and
  Zhang]{xiong2024rlhflowmath}
W.~Xiong, H.~Zhang, N.~Jiang, and T.~Zhang.
\newblock An implementation of generative prm.
\newblock \url{https://github.com/RLHFlow/RLHF-Reward-Modeling}, 2024.

\bibitem[Xiong et~al.(2025)Xiong, Zhao, Yuan, Golovneva, Zhang, Weston, and
  Sukhbaatar]{xiong2025stepwiserstepwisegenerativejudges}
W.~Xiong, W.~Zhao, W.~Yuan, O.~Golovneva, T.~Zhang, J.~Weston, and
  S.~Sukhbaatar.
\newblock Stepwiser: Stepwise generative judges for wiser reasoning, 2025.
\newblock URL \url{https://arxiv.org/abs/2508.19229}.

\bibitem[Yang et~al.(2024)Yang, Zhang, Hui, Gao, Yu, Li, Liu, Tu, Zhou, Lin,
  Lu, Xue, Lin, Liu, Ren, and
  Zhang]{yang2024qwen25mathtechnicalreportmathematical}
A.~Yang, B.~Zhang, B.~Hui, B.~Gao, B.~Yu, C.~Li, D.~Liu, J.~Tu, J.~Zhou,
  J.~Lin, K.~Lu, M.~Xue, R.~Lin, T.~Liu, X.~Ren, and Z.~Zhang.
\newblock Qwen2.5-math technical report: Toward mathematical expert model via
  self-improvement, 2024.
\newblock URL \url{https://arxiv.org/abs/2409.12122}.

\bibitem[Ye et~al.(2025)Ye, Yu, Zhang, Chen, Sadagopan, Huang, Zhang, and
  Beniwal]{ye2025correctnessharmonizingprocessoutcome}
C.~Ye, Z.~Yu, Z.~Zhang, H.~Chen, N.~Sadagopan, J.~Huang, T.~Zhang, and
  A.~Beniwal.
\newblock Beyond correctness: Harmonizing process and outcome rewards through
  rl training, 2025.
\newblock URL \url{https://arxiv.org/abs/2509.03403}.

\bibitem[Zhang et~al.(2026{\natexlab{a}})Zhang, Gao, Ren, and
  Cao]{zhang2026bidirectionalprocessrewardmodel}
L.~Zhang, J.~Gao, X.~Ren, and Z.~Cao.
\newblock The bidirectional process reward model, 2026{\natexlab{a}}.
\newblock URL \url{https://arxiv.org/abs/2508.01682}.

\bibitem[Zhang et~al.(2025{\natexlab{a}})Zhang, Zheng, Wu, Zhang, Lin, Yu, Liu,
  Zhou, and Lin]{prmlessons}
Z.~Zhang, C.~Zheng, Y.~Wu, B.~Zhang, R.~Lin, B.~Yu, D.~Liu, J.~Zhou, and
  J.~Lin.
\newblock The lessons of developing process reward models in mathematical
  reasoning.
\newblock \emph{arXiv preprint arXiv:2501.07301}, 2025{\natexlab{a}}.

\bibitem[Zhang et~al.(2025{\natexlab{b}})Zhang, Zheng, Wu, Zhang, Lin, Yu, Liu,
  Zhou, and Lin]{zhang2025lessonsdevelopingprocessreward}
Z.~Zhang, C.~Zheng, Y.~Wu, B.~Zhang, R.~Lin, B.~Yu, D.~Liu, J.~Zhou, and
  J.~Lin.
\newblock The lessons of developing process reward models in mathematical
  reasoning, 2025{\natexlab{b}}.
\newblock URL \url{https://arxiv.org/abs/2501.07301}.

\bibitem[Zhang et~al.(2026{\natexlab{b}})Zhang, Shan, Song, Li, and
  Ren]{zhang2026linkingprocessoutcomeconditional}
Z.~Zhang, Z.~Shan, K.~Song, Y.~Li, and K.~Ren.
\newblock Linking process to outcome: Conditional reward modeling for llm
  reasoning, 2026{\natexlab{b}}.
\newblock URL \url{https://arxiv.org/abs/2509.26578}.

\bibitem[Zhao et~al.(2025)Zhao, Liu, Zhang, Zhou, Gao, Li, Lyu, Qian, Qi, Li,
  and Zhou]{zhao2025genprmscalingtesttimecompute}
J.~Zhao, R.~Liu, K.~Zhang, Z.~Zhou, J.~Gao, D.~Li, J.~Lyu, Z.~Qian, B.~Qi,
  X.~Li, and B.~Zhou.
\newblock Genprm: Scaling test-time compute of process reward models via
  generative reasoning, 2025.
\newblock URL \url{https://arxiv.org/abs/2504.00891}.

\bibitem[Zheng et~al.(2025)Zheng, Zhang, Zhang, Lin, Lu, Yu, Liu, Zhou, and
  Lin]{zheng2025processbenchidentifyingprocesserrors}
C.~Zheng, Z.~Zhang, B.~Zhang, R.~Lin, K.~Lu, B.~Yu, D.~Liu, J.~Zhou, and
  J.~Lin.
\newblock Processbench: Identifying process errors in mathematical reasoning,
  2025.
\newblock URL \url{https://arxiv.org/abs/2412.06559}.

\bibitem[Zhu et~al.(2025)Zhu, Liu, Li, Tian, Yang, Zhang, Han, Xie, Cui, Zhang,
  Ma, Yu, Ramesh, Wu, Liu, Lu, Zou, and You]{zhu2025llmagentsfaillearn}
K.~Zhu, Z.~Liu, B.~Li, M.~Tian, Y.~Yang, J.~Zhang, P.~Han, Q.~Xie, F.~Cui,
  W.~Zhang, X.~Ma, X.~Yu, G.~Ramesh, J.~Wu, Z.~Liu, P.~Lu, J.~Zou, and J.~You.
\newblock Where llm agents fail and how they can learn from failures, 2025.
\newblock URL \url{https://arxiv.org/abs/2509.25370}.

\bibitem[Zou et~al.(2025)Zou, Yang, Gu, Qiu, Shen, He, and
  Wang]{zou2025reasonfluxprmtrajectoryawareprmslong}
J.~Zou, L.~Yang, J.~Gu, J.~Qiu, K.~Shen, J.~He, and M.~Wang.
\newblock Reasonflux-prm: Trajectory-aware prms for long chain-of-thought
  reasoning in llms, 2025.
\newblock URL \url{https://arxiv.org/abs/2506.18896}.

\end{thebibliography}
\clearpage
\appendix
\section{Benchmark Descriptions}\label{app:benchmarks}

\subsection{Process Supervision Benchmarks}

\textbf{PRMBench}~\citep{song2025prmbenchfinegrainedchallengingbenchmark} is a fine-grained process supervision benchmark that evaluates models on three dimensions: \textbf{Simplicity}, \textbf{Soundness}, and \textbf{Sensitivity}. The benchmark requires identifying both the first erroneous step and all erroneous steps in a reasoning chain, focusing on continuous diagnosis over the entire trajectory.

\textbf{ProcessBench}~\citep{zheng2025processbenchidentifyingprocesserrors} focuses on earliest critical error localization in mathematical reasoning. It evaluates models across four datasets: \textbf{GSM8K}, \textbf{MATH}, \textbf{OlympiadBench}, and \textbf{OmniMath}. The benchmark uses F1 score to measure both error detection and correct step recognition.

\subsection{Cross-Task Generalization Benchmarks}

\textbf{Agent Error Bench}~\citep{zhu2025llmagentsfaillearn} evaluates error localization in LLM-agent trajectories across three environments: \textbf{ALFWorld}, \textbf{WebShop}, and \textbf{GAIA}. The benchmark defines two tasks: \textbf{S} (identifying the earliest critical error step) and \textbf{S+M} (identifying both the error step and the responsible module).

\textbf{PPE (Preference Proxy Evaluations)} evaluates models' ability to make preference judgments between candidate responses. We use the verifiable correctness subset, which includes: \textbf{MMLU-Pro}, \textbf{MATH}, \textbf{GPQA}, \textbf{MBPP-Plus}, and \textbf{IFEval}.

\subsection{Test-Time Search Benchmarks}

For test-time search evaluation, we use six mathematical reasoning benchmarks: \textbf{AIME 2024}, \textbf{AMC 2023}, \textbf{MATH}, \textbf{OlympiadBench}, \textbf{College Math}, and \textbf{Minerva MATH}.

We evaluate under two settings: \textbf{Best-of-8} (sample 8 complete trajectories and select the top-ranked solution) and \textbf{Greedy Guided Search@8} (step-wise 8-candidate search: at each step the policy generates 8 \emph{complete} candidate trajectories, GPRM scores the current step of each under its future trajectory, and a step judged correct is committed before moving on). Because GPRM scores a step from its future trajectory, every candidate step must first be completed into a full trajectory; this completion cost is analyzed in Section~\ref{sec:cost_analysis}.

\section{Data Construction Details}\label{app:data}

\subsection{SFT Data Construction}

We construct the SFT dataset based on \textbf{OpenMathInstruct-2}. The construction process consists of three steps:

\paragraph{Step 1: Seed Sample Selection.}
We select \textbf{80,000} hard samples from the \texttt{train\_1M} subset of OpenMathInstruct-2.

\paragraph{Step 2: Targeted Error Injection.}
For each seed sample $(Q, \mathbf{s})$ where $\mathbf{s}=(s_1,\ldots,s_n)$, we use the teacher model \textbf{Qwen3-235B-Instruct} to inject errors targeting four analysis dimensions:
\begin{enumerate}[leftmargin=*,topsep=0.5pt,itemsep=0pt,parsep=0pt]
    \item \textbf{Forward Consistency} (40K samples)
    \item \textbf{Self Validity} (20K samples)
    \item \textbf{Goal Alignment} (10K samples)
    \item \textbf{Posterior Verification} (10K samples)
\end{enumerate}

For each error type $e \in \mathcal{E}$ and target position $i$, we generate an erroneous step $\tilde{s}_i^{(e)}$ and propagate its effect to subsequent steps, yielding a perturbed trajectory $\tilde{\mathbf{s}}^{(e)}=(s_1,\ldots,s_{i-1},\tilde{s}_i^{(e)},\tilde{s}_{i+1}^{(e)},\ldots,\tilde{s}_n^{(e)})$. The per-dimension counts (40K/20K/10K/10K) are set in proportion to the empirical error-type frequencies estimated from the PRM800K annotations, so that the injection distribution follows the natural occurrence of each diagnostic dimension; a more exhaustive mixture-ratio search is left to future work.

\paragraph{Step 3: Teacher Annotation.}
For each perturbed trajectory, we prompt the teacher model to generate a structured four-dimensional analysis $A_i = (a_i^{\mathrm{F}}, a_i^{\mathrm{P}}, a_i^{\mathrm{G}}, a_i^{\mathrm{S}})$ for the target step $\tilde{s}_i^{(e)}$. If the teacher's initial judgment is inconsistent with the ground-truth label $y_i^*$, we apply self-correction prompting and require re-analysis.

\subsection{GRPO Data Construction}

We construct the GRPO dataset based on \textbf{PRM800K}. The construction process consists of three steps:

\paragraph{Step 1: Difficult Sample Selection.}
We select \textbf{10,000} difficult samples from PRM800K based on three criteria: low confidence, long-horizon reasoning, and delayed-error positions.

\paragraph{Step 2: Future Trajectory Completion.}
For samples where the original trajectory terminates immediately after an error, we use the teacher model to generate the subsequent reasoning process. For an erroneous step $s_i$, we condition on $(Q, \mathbf{s}_{1:i-1}, s_i)$ and generate $M=5$ candidate continuations using temperature $\tau=0.7$ and nucleus sampling ($p=0.9$).

\paragraph{Step 3: Hard-Example Mining.}
We construct the complete global input $X_i^{\text{GRPO}}=(Q,\mathcal{H}_i,s_i,\tilde{\mathcal{F}}_i)$ for each sample. We then perform $K=8$ samplings using the SFT model and retain samples where at least one prediction disagrees with the gold label. This yields approximately \textbf{20,000} GRPO training instances.

\subsection{Generalization to General Reasoning Tasks}
\label{sec:ppe_generalization}

To assess whether GPRM's capabilities extend beyond mathematical reasoning, we evaluate it on the verifiable correctness subset of the PPE (Process-level Preference Evaluation) benchmark~\citep{ppe_citation}, which covers diverse reasoning domains including multi-choice QA (MMLU-Pro), mathematics (MATH), science reasoning (GPQA), code generation (MBPP-Plus), and instruction following (IFEval).

As shown in Table~\ref{tab:ppe_results}, GPRM-4B achieves a mean score of 0.64, outperforming GPT-4o-mini (ArenaHard) (0.63) and Llama-3.1-70B (ArenaHard) (0.63), despite being significantly smaller in size. This performance approaches Claude 3.5 (ArenaHard) (0.68) and Athene-RM-70B (0.69), demonstrating that GPRM learns general trajectory evaluation capabilities that transfer effectively beyond mathematical domains. Notably, GPRM-4B shows strong performance across diverse tasks, achieving 0.71 on MATH, 0.67 on MMLU-Pro, 0.65 on MBPP-Plus, and 0.58 on GPQA, confirming its robustness as a general-purpose process reward model.

\begin{table}[tbp]
\centering
\caption{Performance on the \textbf{verifiable correctness subset} of \textbf{PPE}. $\star$ denotes our model.}
\label{tab:ppe_results}
\small
\setlength{\tabcolsep}{6pt}
\renewcommand{\arraystretch}{1.1}
\begin{tabular}{lcccccc}
\toprule
\textbf{Reward Model} & \textbf{MMLU-Pro} & \textbf{MATH} & \textbf{GPQA} & \textbf{MBPP-Plus} & \textbf{IFEval} & \textbf{Mean} \\
\midrule
Athene-RM-70B                    & 0.77 & 0.79 & 0.59 & 0.68 & 0.62 & \textbf{0.69} \\
Claude 3.5 (ArenaHard)$^{\dagger}$ & \textbf{0.81} & \textbf{0.86} & \textbf{0.63} & 0.54 & 0.58 & 0.68 \\
Llama-3-OffsetBias-RM-8B         & 0.62 & 0.68 & 0.55 & \textbf{0.74} & 0.62 & 0.64 \\
GPT-4o-mini (ArenaHard)$^{\dagger}$ & 0.71 & 0.81 & 0.57 & 0.54 & 0.56 & 0.63 \\
Llama-3.1-70B (ArenaHard)$^{\dagger}$ & 0.73 & 0.73 & 0.56 & 0.58 & 0.56 & 0.63 \\
internLM2-20B-Reward             & 0.68 & 0.70 & 0.57 & 0.58 & 0.62 & 0.63 \\
Athene-RM-8B                     & 0.68 & 0.71 & 0.55 & 0.62 & 0.57 & 0.62 \\
ArmoRM-Llama3-8B-v0.1            & 0.66 & 0.71 & 0.57 & 0.54 & 0.58 & 0.61 \\
Skywork-Reward-Llama-3.1-8B      & 0.64 & 0.70 & 0.57 & 0.52 & 0.61 & 0.61 \\
Nemotron-4-340B-Reward           & 0.70 & 0.65 & 0.57 & 0.49 & 0.63 & 0.61 \\
internLM2-7B-Reward              & 0.67 & 0.73 & 0.55 & 0.44 & \textbf{0.64} & 0.60 \\
Llama-3.1-70B (Alpaca)$^{\dagger}$ & 0.66 & 0.66 & 0.56 & 0.52 & 0.56 & 0.59 \\
Claude 3.5 (Alpaca)$^{\dagger}$    & 0.66 & 0.63 & 0.56 & 0.52 & 0.57 & 0.59 \\
Skywork-Reward-Gemma-2-27B       & 0.54 & 0.63 & 0.53 & 0.59 & 0.54 & 0.56 \\
GPT-4o-mini (Alpaca)$^{\dagger}$   & 0.57 & 0.64 & 0.53 & 0.52 & 0.56 & 0.56 \\
NaiveVerbosityModel              & 0.48 & 0.50 & 0.48 & 0.31 & 0.52 & 0.46 \\
\midrule
\rowcolor{gray!15}
$\star$ GPRM-4B                  & 0.67 & 0.71 & 0.58 & 0.65 & 0.59 & 0.64 \\
\bottomrule
\end{tabular}
\end{table}

\section{Training Details}\label{app:training}

\subsection{SFT Training}

\textbf{Model Architecture.} We initialize GPRM from \textbf{Qwen3-4B-Instruct} and \textbf{Qwen3-14B}.

\textbf{Training Hyperparameters.}
\begin{itemize}[leftmargin=*,topsep=0.5pt,itemsep=0pt,parsep=0pt]
    \item Learning rate: $2 \times 10^{-5}$ with cosine decay
    \item Batch size: 128
    \item Training epochs: 3
    \item Optimizer: AdamW with $\beta_1=0.9$, $\beta_2=0.95$
    \item Weight decay: 0.1
    \item Gradient clipping: 1.0
    \item Warmup steps: 500
\end{itemize}

\textbf{Training Objective.} We use the standard autoregressive maximum likelihood objective:
\[
\mathcal{L}_{\text{SFT}}(\phi) = -\sum_{(X_i^{\text{SFT}},A_i)\in\mathcal{D}_{\text{SFT}}} \log P_\phi(A_i \mid X_i^{\text{SFT}}).
\]

\subsection{GRPO Training}

\textbf{Training Hyperparameters.}
\begin{itemize}[leftmargin=*,topsep=0.5pt,itemsep=0pt,parsep=0pt]
    \item Learning rate: $1 \times 10^{-6}$ with constant schedule
    \item Batch size: 64
    \item Training epochs: 2
    \item Optimizer: AdamW with $\beta_1=0.9$, $\beta_2=0.95$
    \item Weight decay: 0.01
    \item Gradient clipping: 1.0
    \item Number of samples per input: $K=8$
    \item Clipping threshold: $\epsilon=0.2$
\end{itemize}

\textbf{Reward Function.} We use a binary outcome-based reward:
\[
r(\hat{A}_i^{(k)}) = \mathbb{I}[\operatorname{Parse}(\hat{A}_i^{(k)})=y_i^*].
\]

\textbf{GRPO Objective.} The optimization objective is:
\begin{multline*}
\mathcal{L}_{\text{GRPO}}(\phi) = -\mathbb{E}_{X_i^{\text{GRPO}}\sim \mathcal{D}_{\text{hard}}}
\Bigg[
\frac{1}{K}\sum_{k=1}^{K}
\min\Big(
\rho_i^{(k)}\operatorname{Adv}_i^{(k)}, \\
\operatorname{clip}(\rho_i^{(k)},1-\epsilon,1+\epsilon)\operatorname{Adv}_i^{(k)}
\Big)
\Bigg],
\end{multline*}
where $\rho_i^{(k)} = \pi_\phi(\hat{A}_i^{(k)}\mid X_i^{\text{GRPO}}) / \pi_{\phi_{\text{ref}}}(\hat{A}_i^{(k)}\mid X_i^{\text{GRPO}})$ is the importance sampling ratio, and $\operatorname{Adv}_i^{(k)} = r_i^{(k)}-\frac{1}{K}\sum_{j=1}^{K} r_i^{(j)}$ is the group-relative advantage function.

\subsection{Computational Resources}

All experiments are conducted on NVIDIA A100 80GB GPUs. Training times:
\begin{itemize}[leftmargin=*,topsep=0.5pt,itemsep=0pt,parsep=0pt]
    \item \textbf{GPRM-4B SFT}: 12 hours on 8 GPUs
    \item \textbf{GPRM-4B GRPO}: 8 hours on 8 GPUs
    \item \textbf{GPRM-14B SFT}: 36 hours on 16 GPUs
    \item \textbf{GPRM-14B GRPO}: 24 hours on 16 GPUs
\end{itemize}

\section{Evaluation Protocols}\label{app:eval}

\subsection{Process Supervision Evaluation}

\textbf{PRMBench.} We evaluate on all three dimensions and report the overall average score. For each problem, the model predicts step-level correctness labels, and we compute accuracy by comparing predictions with gold annotations.

\textbf{ProcessBench.} We report F1 scores for both error detection and correct step recognition on each of the four datasets. The average F1 is computed as the unweighted mean across all datasets.

\subsection{Cross-Task Generalization Evaluation}

\textbf{Agent Error Bench.} We evaluate on two tasks: \textbf{S} (accuracy of identifying the earliest critical error step) and \textbf{S+M} (accuracy of jointly identifying the error step and the responsible module). Results are reported separately for each environment and averaged.

\textbf{PPE.} For each prompt with multiple candidate responses, we apply GPRM to predict step-level correctness for each candidate trajectory, compute a trajectory score as the ratio of correct steps to total steps, and select the candidate with the highest score. We report the accuracy of preference judgments.

\subsection{Test-Time Search Evaluation}

\textbf{Best-of-8.} For each problem, we sample 8 complete trajectories, apply GPRM to compute a trajectory score for each candidate, and select the trajectory with the highest score. We report the accuracy of final answers.

\textbf{Greedy Guided Search@8.} For each problem, at each step the policy first generates 8 \emph{complete} candidate trajectories that share the committed prefix but differ in how the current step continues into the future. GPRM then scores the current step of each candidate jointly with its future trajectory $\mathcal{F}_i$, and a step judged correct (one of the candidates scoring CORRECT) is committed to extend the trajectory; the committed step plus its prefix is fed back to the policy, and the next step is searched in the same way. The process repeats until the model emits a final answer. We report the accuracy of final answers. Because scoring a candidate step requires its future trajectory, the policy must complete each candidate before GPRM can evaluate it; the resulting compute overhead is analyzed in Section~\ref{sec:cost_analysis}.

\end{document}